# Application of Natural Language Processing to Determine User Satisfaction in Public Services

Radoslaw Kowalski
University College London
radoslaw.kowalski.14@ucl.ac.uk

Marc Esteve
University College London and ESADE
marc.esteve@ucl.ac.uk

Slava J. Mikhaylov
University of Essex
s.mikhaylov@essex.ac.uk

Abstract

Research on customer satisfaction has increased substantially in recent years. However, the relative importance and relationships between different determinants of satisfaction remains uncertain. Moreover, quantitative studies to date tend to test for significance of pre-determined factors thought to have an influence with no scalable means to identify other causes of user satisfaction. The gaps in knowledge make it difficult to use available knowledge on user preference for public service improvement. Meanwhile, digital technology development has enabled new methods to collect user feedback, for example through online forums where users can comment freely on their experience. New tools are needed to analyze large volumes of such feedback. Use of topic models is proposed as a feasible solution to aggregate open-ended user opinions that can be easily deployed in the public sector. Generated insights can contribute to a more inclusive decision-making process in public service provision. This novel methodological approach is applied to a case of service reviews of publicly-funded primary care practices in England. Findings from the analysis of 145,000 reviews covering almost 7,700 primary care centers indicate that the quality of interactions with staff and bureaucratic exigencies are the key issues driving user satisfaction across England.

# Introduction

User satisfaction is a litmus test of public service effectiveness (Lavertu 2014). How citizens value public services may differ from what organizational experts and decision-makers would understand as tokens of good performance (Lavertu 2014; Sanders and Canel 2015). It is therefore important to effectively examine the determinants of user satisfaction so that managers of public organizations comprehend and acknowledge the user perspective on public service quality. A robust understanding of public preferences helps to ensure that managers of public institutions make decisions that are aligned with the public need.

At the same time, digital technologies have led to the creation of a host of new opportunities for the collection of service user feedback (Bauer and Nanopoulos 2014; Kong and Song 2016). On the one hand, these new data resources can be very insightful because they contain much more detail about what makes citizens (un)happy with public services compared to traditional survey methods. They are widely utilized for this reason in private sector companies (see, for example, Qi et al. 2016), although so far with scant examples of good practice in the public sector (Hogenboom et al. 2016). On the other hand, however, there are problems with using these new data resources. First, they can be too large to read and analyze manually (Kong and Song 2016). Second, the obtainable data may predominantly consist of unstructured text which is hard to summarize with standard statistical techniques (Kong and Song 2016). Finally, it is often difficult to pinpoint the sample biases because the identities of authors are uncertain (Yang 2010). The volume and structure of text feedback, e.g. in the form of reviews, makes it difficult to use it to obtain operationally useful information about the causes of user satisfaction from public services. Simultaneously, existing tools developed for private organizations may not be adequate for use in the public sector. Public organizations require insights into service user preferences in a situation where citizens are "forced customers" (Di Pietro, Guglielmetti Mugion, and Renzi 2013) and where public organizations, such as schools or hospitals, must fulfill objectives which may be unrelated to service demand or profitability (Brownson et al. 2012).

This study addresses the shortages in user satisfaction understanding through an analysis of unstructured text feedback. Unstructured and anonymous feedback can help provide a substantial answer to the research question: "What are the determinants of user satisfaction in public services?". Large quantities of unstructured feedback can be summarized with natural language processing (NLP) models such as topic models in order to obtain actionable insights (Blei, Ng, and Jordan 2003; Hogenboom et al. 2016). Furthermore, the insights from topic modeling can be compared against other analyses such as surveys to systematically evaluate the validity and reliability of text-derived insights. The article offers two contributions to the public management field: 1) it evaluates a comprehensive model of determinants of user satisfaction from public services, and 2) offers an effective method to analyze big data for public services. The contributions stem from the implementation of NLP to solve a public management analytical problem.

## User Satisfaction with Public Services

User satisfaction is very important for public sector organizations (Andrews, Boyne, and Walker 2011; James 2009; Kelly 2005). Public service organizations should pay attention to it because public satisfaction relates to the levels of trust citizens have in political leadership and government institutions (Christensen and Laegreid 2005; Ellis 2015; Kampen, Van de Walle, and Bouckaert 2006). Public institutions ought to also strive to get things right each time for service users because their satisfaction can drop quickly when services fail and may rise only marginally when the experience is positive (Kampen et al. 2006). Governments are more legitimate when governance arrangements are fit to solve the issues faced by the public (Fung 2015; James and Van Ryzin 2015; Potapchuk 2016). Lack of satisfaction from government services may lead to citizens not cooperating with the government and to political instability (Córdova and Layton 2016; Schofield and Reeves 2015). Furthermore, user satisfaction is also important in the process of public service delivery (Di Pietro, Guglielmetti Mugion, and Renzi 2013). Public organizations tend to be more aligned with the interests of end users when they collect and act upon user feedback (Beeri and Yuval 2013; Park 2015). The feedback of users may be useful not only in allocating extra funds, but also for rethinking funding priorities in times of economic crisis (Jimenez 2013). Heeding user feedback may result in better staff-client relations and higher job satisfaction on the part of frontline workers (Brown and Calnan 2016; Franco-Santos, Lucianetti, and Bourne 2012). It is also uniquely useful for including the personal aspect of the service into performance evaluations (Nash 2015) and helps with making meaningful comparisons between providers who score very similarly in standard performance measures (Alemi et al. 2012; James, Calderon, and Cook 2017). Furthermore, organizational reforms that take into account user feedback can lead to increased organizational efficiency (Jimenez 2013), for instance leading to higher treatment rates in relation to preventable diseases (Brown and Calnan 2016; Kiernan and Buggy 2015; Poku 2016). Studies point to many successful examples wherein engagement with citizens regarding decisions taken about public services can be fiscally sustainable (Beeri and Yuval 2013; Orlitzky, Schmidt, and Rynes 2003; Park 2015; Rahman and Bullock 2005; Yuk-ping 2015), and may even be beneficial for the re-election chances of political leaders (Park 2014).

The importance of user satisfaction in public administration led to many attempts to refine its measurement (Andrews et al. 2007; Andrews et al. 2011; Beaussier et al. 2015). End user data collection is difficult to substitute because service users themselves tend to adopt the predominant views held within public organizations once they are regularly involved in the decision-making process, especially when citizen involvement is not perceived by them as decisive in resource allocation (Greer et al. 2014; Grohs, Adam, and Knill 2015; Rutherford and Meier 2015). Frequently, end user satisfaction is estimated with proxy values and included in the performance measurement system of public organizations (e.g., Brenes, Madrigal, and Requena 2011; Grigoroudis, Orfanoudaki, and Zopounidis 2012; Gunasekaran and Kobu 2007; Kelman and Friedman 2009). Targets, such as service speed, may substitute citizen opinions even when there may be little connection between self-reported client concerns and estimations of their satisfaction. In consequence, seemingly data-driven and transparent performance evaluations are biased with falsely positive performance scores assigned to assessed organizations (Andersen, Heinesen, and Pedersen 2016; Bischoff and Blaeschke 2016; Lowe and Wilson 2015; Rutherford and Meier 2015). The structure of performance evaluations incentivizes resource allocation towards measured dimensions of service quality regardless of whether it benefits service users (Brown and Calnan 2016; Farris et al. 2011; Gao 2015; Hood and Dixon 2013; Lowe and Wilson 2015; Poku 2016).

While the importance of user satisfaction to improve organizations is widely acknowledged, the existing literature suggests that there is no consensus among scholars and practitioners over how to include user feedback in the organizational performance evaluation process. Available studies tend to fall into two broad categories. The first includes proponents of evidence-based policy making and New Public Management (NPM): researchers who adopt an ontological assumption that it is possible to attain a single and fairly static performance evaluation system that is superior to reliance on sets of discrete and sometimes contradictory viewpoints (Head 2016; Isett, Head, and Vanlandingham 2016; Kelman and Friedman 2009; Osborne, Radnor, and Nasi 2012; Tucker 2004). In other words, it is assumed that some or other form of rationality is attainable for the benefit of both individuals and society. Given this assumption, authors tend to argue that user feedback is a biased source of information from self-interested individuals whose perceptions can be manipulated according to the service information given (Im et al. 2012; Jensen and Andersen 2015; Ma 2017; Marvel 2016; Moon 2015; Moynihan, Herd, and Harvey 2014). Those authors tend to imply that the perspective of researchers on organizational performance is value-neutral and selfless, mostly in contrast to service users (Head 2016; Pisano 2016). That said, some researchers argue that user satisfaction should form the core influence in performance evaluations of public services, because it is an enhanced resource for the construction of "universal rationality" compared to other perspectives (Osborne, Radnor, and Nasi 2012). This family of studies tends to place emphasis on the suppression of rejected subjectivities in the performance evaluation process for the sake of efficiency (Head 2016; Jensen and Andersen 2015; Moon 2015; Osborne, Radnor, and Nasi 2012), while at the same time increasing the amount of data processed for performance evaluations to make them more robust (Boswell 2015; Dickinson and Sullivan 2014; Head 2016; Lavertu 2014; Ma 2017). Furthermore, measurement transparency is also considered critical for obtaining better informed (i.e. assumed as "closer to being objective") end user feedback (Ho and Cho 2016; Larrick 2017; Michener and Ritter 2017). This approach to understanding organizational performance is sometimes confirmed through success stories where evidence-based policy making, that has ignored the voice of end users, has led to improvements in organizational performance (e.g., Kelman and Friedman 2009). However, most reviewed studies in favor of NPM do not point to concrete examples where evidence-based performance measurement resulted in meaningful quality improvements. This observation is in line with wider findings that NPM has not led to major improvements in public service organizations (Hood and Dixon 2015a, 1–19; Reay, Berta, and Kohn 2009).

In contrast to the supporters of evidence-based policy making, the second family of studies on use of end user feedback in public organizations includes arguments that point to the empirical failures of NPM. The corollary of the critiques tends to be an implicit (Bevan and Hood 2006; Hood and Dixon 2015a, 1–19; Pflueger 2015) or cautiously explicit (Amirkhanyan, Kim, and Lambright 2013; Liu 2016; O'Malley 2014) assumption that what constitutes organizational performance is not static, but rather evolves through deliberation between interested parties, each of which has a limited and shifting understanding of what constitutes public service effectiveness (Liu 2016). Given this understanding of organizational performance, multiple interacting perspectives on public service are assumed to lead to superior outcomes compared to a single, static perspective on performance (Liu 2016). End user feedback is valued as an important element of the continuous performance refinement process of public services (Amirkhanyan, Kim, and Lambright 2013; Andersen, Heinesen, and Pedersen 2016). Moreover, critics of NPM argue that a singular perspective on organizational performance itself represents a subjective understanding of what constitutes service quality (DeBenedetto 2017; Rabovsky 2014) and tends to marginalize the voice of service users within the organizational objective-setting process (Amirkhanyan, Kim, and Lambright 2013; DeBenedetto 2017; Kroll 2017; Larrick 2017; Lavertu 2014; Worthy 2015). Voices of more influential individuals and pressure groups dominate organizational priorities when NPM approach is practiced (Worthy 2015). In effect, it appears that increased accountability in line with the principles of NPM lowered public satisfaction in terms of government services (Hood and Dixon 2015b, 265–267; Lavertu 2014; Tucker 2004) and was detrimental to the quality of the democratic process (James and Moseley 2014; Van Loon 2017). As a result, citizens who become more skeptical of their own ability to influence how public services are provided give up on voicing dissatisfaction and resort to development of game-playing skills to bargain with, conspire against, and deceive public institutions processes (James and Moseley 2014; Van Loon 2017). Furthermore, the service quality issues faced by citizens are increased by the inherent weaknesses of any transparent and evidence-based performance improvement system (Amirkhanyan,

Kim, and Lambright 2013; Bernstein 2012; Brown and Calnan 2016; Gao 2015; Johannson 2016; Lavertu 2014; Poku 2016; Worthy 2015). Critiques of NPM suggest that accurate measurement of complex service quality phenomena is impossible to achieve with short lists of static and crude performance metrics (Bernstein 2012; Brown and Calnan 2016; DeBenedetto 2017; Gao 2015; Johannson 2016; Lavertu 2014; Ma 2017; Poku 2016; Rabovsky 2014). Moreover, the NPM-style performance measurement process cannot become very complex because decision-makers themselves cease to find the measures useful (Lavertu 2014), and possibly also because the cost of making additional metrics can be prohibitive.

Evidence-based policy-making does not seem to be an ideal solution for organizing public services to ensure end user satisfaction. The voice of citizens should be included in the process of setting up services, rather than crowded out; but the real question is how this can be achieved. The inclusion of the service user voice in decisions requires a robust understanding of how and why they are satisfied. Citizen satisfaction is known to correlate (but often non-linearly) with their socio-economic status, education, and employment history (Christensen and Laegreid 2005; Harding 2012; Mcguire et al. 2014; Yang 2010), demographic background (Yang 2010), and relevant information to which they have been exposed and understood in their own way (Hong 2015; Im et al. 2012; James and Moseley 2014; Lavertu 2014; Mason, Baker, and Donaldson 2011; Villegas 2017). Citizens' opinions also tend to have little to do with the formal measures of organizational performance used for assessments within organizations (Harding 2012; Ma 2017; Moynihan, Herd, and Harvey 2014; Sanders and Canel 2015; Voutilainen 2016) and the opinions of organizational managers (Andersen and Hjortskov 2016; Sanders and Canel 2015), but they are simultaneously related to how frontline public workers experience service provision (Amirkhanyan, Kim, and Lambright 2013; Raleigh et al. 2009). The satisfaction of citizens is also related to how, if at all, they use the commented-upon public services (Brown 2007; Im et al. 2012; Ladhari and Rigaux-bricmont 2013; Lopez et al. 2012; Pierre and Røiseland 2016; Van Ryzin and Charbonneau 2010), and in what way they are involved with their provision (Larsen and Blair 2010; Sanders and Canel 2015; Scott and Vitartas 2008; Taylor 2015). For example, Taylor (2015) has shown how citizens who are aware of their income tax money contributing to specific services tend to place greater scrutiny on those services compared to those with little fiscal contribution to the same services.

Users also strive to express their satisfaction of service experience in a way that is consistent with their underlying knowledge, beliefs, and opinions (Barrows et al. 2016; Brown 2007; Harding 2012; Ladhari and Rigaux-bricmont 2013) as well as emotional attachments (Fledderus 2015; Gee 2017; Lawton and Macaulay 2013; Ma 2017). At the same time, those opinions may not always be thought through due to the limited time available and cognitive capacity to develop a belief (Andersen and Hjortskov 2016; Barrows et al. 2016; Etzioni 2014; Villegas 2017). Furthermore, the reported service satisfaction may relate to the motivation of individuals to provide feedback which may be loosely related to their actual service experience (Brenninkmeijer 2016; Duit 2016; Pauna and Luminita 2015; Potapchuk 2016; Sanders and Canel 2015; Zieliński 2016). For example, Brenninkmeijer (2016) reports that in one study carried out in the 2010s in the Netherlands, over 70% of offenders caught on traffic offences by police were positive about the performance of police services.

While researchers have uncovered multiple determinants of user satisfaction from public services, it often remains unclear how those determinants relate to one another, and whether the interactions between determinants are the same irrespective of context and the passage of time. Moreover, it is often unclear whether the aspects of user satisfaction that researchers and/or commissioners of research choose to investigate constitute a complete list of determinants (Lavertu 2014; Roberts et al. 2014). Factors outside the scope of already well-known determinants of satisfaction may bias insights from commissioned studies in unpredictable ways, and the avenues of how and why it happens are often entirely unclear (Pierre and Røiseland 2016). Similarly, researchers of user satisfaction from public services test a wide range of theories about what determines it, which makes it difficult to construct a robust, holistic understanding of what matters most to service users and why. For example, they may focus on investigating the impacts of available information (James and Moseley 2014; Marvel 2016), self-centered utility maximization (Jensen and Andersen 2015), emotions (Ladhari and Rigaux-bricmont 2013), sense of identity (Mcguire et al. 2014), unconscious tendency towards conformity (Sanders and Canel 2015), or the level of physical involvement with the services under review (Loeffler 2016). As a result, it is not certain why official performance metric achievement is often incongruent with citizens' satisfaction levels (Brenninkmeijer 2016) or how and why citizens are satisfied in their consumption of public e-services (Im et al. 2012). Furthermore, narratives used by citizens to explain their (dis)satisfaction may be unknown even when behavior is well understood (Müssener et al. 2016). The available literature indicates that there is a gap in understanding the relative importance and relationships between the determinants of service user satisfaction, as well as an absence of understanding as to whether some factors influencing user satisfaction are omitted or misrepresent how service users form their satisfaction evaluations.

## User Feedback as a Measure of Satisfaction

Systematic understanding of the determinants of citizen satisfaction in terms of public services requires a robust means of capturing this voice in a manner that is appropriate in a given context. As mentioned above, the effective inclusion of user perspective in decisions about public services can help to build satisfaction of public institutions (Fung 2015). It is also desired as a prerequisite of successful democratic governance (Feldman 2014; Fung 2015) and is a necessary means to solve pressing problems regarding service output performance (Fung 2015; Mahmoud and Hinson 2012; Richter and Cornford 2007). Physical participation of citizens in public decision-making is one way for public authorities to engage and understand the service user perception of public services (Fung 2015). The approach can be successful at bringing meaningful change to institutions that benefits users and increases their satisfaction of public services (Moon 2015). However, at the same time, the public's direct participation in decisions is not easy to scale to more complex problems within public services. In an applied context, it may also politicize otherwise quick administrative decisions with poor marginal returns for the additional effort put into the decision (Bartenberger and Sześciło 2016). Moreover, in many institutional contexts it is difficult to capture enough interest from service users to keep them regularly involved in decision-making (Fung 2015; Greer et al. 2014). Liu (2016) argues, with hands-on examples, that the understanding of service user preferences could improve with information technologies and lead to new modes of decision-making.

The representation of the service user voice through data collection and summarization is an alternative to direct citizen participation in situations where the latter is not feasible. Surveys are a widely-used tool to measure user satisfaction with the quality of public services and help to keep providers on track (Van de Walle and Van Ryzin 2011). At times, experiments and/or qualitative research is also carried out to help explore the strengths and weaknesses of implemented methods of capturing the public voice (e.g., James and Moseley 2014; Mahmoud and Hinson 2012; Richter and Cornford 2007). Those research methods, unlike surveys, tend to be one-off with the aim of understanding specific problems with public services. The high running costs involved may be among the reasons why reviewed studies rarely mention the use of experiments or qualitative research approaches for the day-to-day inclusion of the public's voice in decisions about public services. At the same time, popular user surveys also face limitations in measuring user satisfaction (Olsen 2015). There are no established tools to update the survey satisfaction measurement to changing conditions (Burton 2012; Madsen et al. 2015; Schofield and Reeves 2015). Inability to carry out frequent surveys also makes them less useful for daily monitoring of service satisfaction, for example to observe in real-time the impact of organizational change (Burton 2012; Gee 2017; Walker and Boyne 2009). Furthermore, feedback received through restricted lists of survey questions tends to oversimplify the reasons for user satisfaction (Amirkhanyan, Kim, and Lambright 2013; Mcguire et al. 2014) and may be biased by survey structure (Gee 2017; Schofield and Reeves 2015; Van de Walle and Van Ryzin 2011). The final survey outputs may also be less useful where user satisfaction scores are similar between many public service providers (Voutilainen et al. 2015). Therefore, both practitioners and academics encourage the introduction of other forms of data beyond surveys to more effectively gauge the determinants of user satisfaction regarding public services (Amirkhanyan, Kim, and Lambright 2013; Andersen, Heinesen, and Pedersen 2016; Brenninkmeijer 2016; Lavertu 2014).

The limitations of existing methods to meaningfully represent the determinants of end user satisfaction encourage the search for alternative methods of feedback data collection. First, alternative forms of user satisfaction measurement should adopt a view that the meaning of organizational performance changes dynamically and depends upon what end users, as well as other relevant individuals such as political decision-makers and public servants, think. This performance conceptualization can help to avoid the reproduction of deficiencies in evidence-based policy making. Those deficiencies include the suppression of the less powerful voice of service users within the performance measurement process (Brown and Calnan 2016; O'Leary 2016) and the measurement of user satisfaction with methods that quickly lose their relevance, requiring effort to develop a replacement (Gao 2015; Johannson 2016). Alternative data resources have the potential to help improve public services, including through new types of interactions with citizens (Rogge, Agasisti, and De Witte 2017). New types of data, such as network signals and written feedback, have already proved their usefulness in service improvements such as e-government, traffic control, and crime detection (Rogge, Agasisti, and De Witte 2017). At the same time, the new technological possibilities require further effort in order to utilize new data within the public policy domain. The sheer volume of data is challenging to handle (Grimmer and Stewart 2013) and decision-makers may not be fully able to collect, process, visualize, and interpret them (Brenninkmeijer 2016; Hogenboom et al. 2014; Lavertu 2014; Rogge, Agasisti, and De Witte 2017). Furthermore, public policy researchers highlight the ethical issues inherent in handling personal data, including respect for individual privacy and security as well as concerns around the quality of democratic processes (O'Leary 2016). The tools developed to handle complex data from service users should be designed with the intention to address those concerns while offering added value to the quality of public services.

Written reviews of public services are one data resource that can be very relevant in capturing the voice of service users and including it in the public decision-making process. Online written reviews can help to address the issues of privacy since they can be posted anonymously. At the same time, anonymous online reviews may still be a valid resource for decision-makers within public institutions, despite their uncertain sample and complex model biases (Grimmer and Stewart 2013). This is because they can be validated against state-of-the-art structured forms of user feedback, such as carefully drafted surveys with large numbers of reviewers (Grimmer and Stewart 2013; Liu et al. 2017; Rogge, Agasisti, and De Witte 2017). Furthermore, the requirements of basic literacy in any language combined with access to the internet can make online forums a channel wherein almost every public service user could contribute and inform research and practice, as well as source information that may prove useful to them. The ease of use of online forums results in written reviews being a potential means for ensuring the equitable distribution of services (Kroll 2017), and for addressing concerns about the quality of the democratic decision-making process (O'Leary 2016). Moreover, organizations assessed based on user review content may be relatively less able to manipulate performance scores in ways that reduce user satisfaction (Hood and Dixon 2015b, 265–267). In addition, the likelihood of decision-makers making poor decisions due to over-reliance on very narrow understandings of service quality is reduced (Luciana 2013; Pflueger 2015). Thus, online reviews could be helpful in understanding and including citizens in decisions about how to provide public services, especially in cases when the public's physical participation in decisions would be unproductive, or when opportunities to participate would unlikely be engaging enough to maintain public involvement.

A key challenge in using online written reviews for inclusive public policy and research is how to process them in a way that is scalable and meaningful for public decision-makers. Fortunately, unsupervised machine learning models, such as topic models, are already well known to simplify insights from written reviews into relatively straightforward numeric summaries in near real-time and regardless of their quantity (Bannister 2015; Blei, Ng, and Jordan 2003; Griffiths and Steyvers 2004; Nguyen and Shirai 2015; Yang et al. 2012). An advantage of these over user surveys is that they can automatically adapt to changes in how and about what users write (Blei and Lafferty 2006; Dai and Storkey 2015) without prior assumptions or constraints about which aspects of a service reviewers can express their satisfaction (Blei, Ng, and Jordan 2003). Several studies have attempted an analysis of written user feedback from services using machine learning algorithms for organizational improvement (Gao and Yu 2016; Gray 2015; Peleja et al. 2013; Rogge, Agasisti, and De Witte 2017; Sharma et al. 2016). However, none has established a firm relationship as to how key themes identified in online written reviews with topic modeling relate to established measures of user satisfaction, such as satisfaction surveys. The knowledge gap must be filled before online written reviews can be used reliably as a measure of user satisfaction that supports the provision of public services (Gee 2017; Grimmer and Stewart 2013; Rogge, Agasisti, and De Witte 2017). Furthermore, the relationship between survey outcomes and the content of written reviews can help researchers to understand how reviewer narratives relate to numerically expressed satisfaction with public services on the dimensions included in the survey.

## Data

In the present article, the evaluation of the link between satisfaction surveys and unstructured reviews is carried out on a dataset of online reviews about publicly funded primary care (GP) services in England. Reviews were downloaded in .xml format from a dedicated service provided by NHS Choices, an NHS organization responsible for handling feedback data[1], and transformed into a .csv table format used for modeling with R programming language. The downloaded online reviews were posted from July 2013 to January 2017, covering almost 7,700 GP practices. Completed reviews used in this study constitute 145,000 (about 89% of all reviews).

The reviews corpus was pre-processed following the standard in the field (Grimmer and Stewart 2013). We lowercased and stemmed the tokens; and removed numbers, punctuation, stop words, tokens shorter than three characters, and tokens that appeared fewer than 10 times and more than 100,000 times in the corpus. Pre-processing removed 37,708 terms that occurred 77,976 times in GP reviews. The final corpus contained 7,660 terms that occurred over 6 million times in the dataset.

Each month, anonymous users posted between 3,000 and 5,000 written comments accompanied by 5-point Likert-scale star ratings of six aspects of their GP service experience. The Likert-scale star ratings related to survey statements: 1) "Are you able to get through to the surgery by telephone?", 2) "Are you able to get an appointment when you want one?", 3) "Do the staff treat you with dignity and respect?", 4) "Does the surgery involve you in decisions about your care and treatment?", 5) "How likely are you to recommend this GP surgery to friends and family if they needed similar care or treatment?", and 6) "This GP practice provides accurate and up to date information on services and opening hours". The reviews were 5-6 sentences long on average, with a median length of 5 sentences.

It should be noted that there are no socio-demographic attributes for users posting the data, so the sample could be skewed towards some demographic. However, on qualitatively reading through the reviews that seems unlikely. Moreover, anyone can comment on the website and evaluate GP practices. That said, NHS Choices administrators manually remove malicious messages from the server.

---

[1] See more about NHS Choices at:
http://www.nhs.uk/aboutNHSChoices/aboutnhschoices/Pages/what-we-do.aspx, viewed on 17 September 2017

Furthermore, NHS Choices staff ensure that unfavorable but legitimate reviews remain consistently in the dataset across England[2].

## Topic Modeling

The written comments of users were modeled with LDA (Latent Dirichlet Allocation) topic model implemented with *stm* software package for R programming language[3]. Topic models are tools for organizing a collection of written documents, for instance forum posts, open-ended survey responses or formal speeches, into several key themes. For example, some topics derived from reviews in this study may be about thanking doctors, complaining about reception staff, or commenting about the quality of GP facilities. Topic modeling is especially useful for analyzing written documents when manual labeling of documents is not feasible due to their high volume, and when new documents are continually being added to the dataset and require processing.

Following Blei (2012), LDA assumes that all documents in the corpus share the same set of topics, and each topic is a random probability distribution over the words present in documents of the dataset. The algorithm begins calculations by giving a random allocation of topics to every document in the corpus. Next, for each word in every document, the algorithm first picks a topic from the random distribution topics of a given document and then picks a word from the selected distribution over words (i.e. the selected topic).

The model requires human input in setting the number of topics to uncover within the dataset. We follow Roberts et al. (2015) and select the optimal number of topics as a balance between exclusivity and semantic coherence. Our analysis shows that 57 topics is the optimal setting for our data. The supplementary materials discuss the selection process in more detail.

---

[2] See for further details
http://www.nhs.uk/aboutNHSChoices/aboutnhschoices/termsandconditions/Pages/commentspolicy.aspx
[3] Further details about the *stm* software library used in R programming language for implementation of the model is available at: https://CRAN.R-project.org/package=stm, viewed on 17 September 2017

Once the topic model has been estimated, the meaning of each can be deduced from words appearing with the highest probability within that topic. This is often completed by human labeling. Table 1 presents the labels for 57 estimated topics from our data. In the supplementary material, we provide full details on the topic labeling exercise.

The key themes extracted from text reviews with the topic model relate to a range of patient experiences, both positive and negative. Most topics point to concrete aspects of public GP service, such as availability of parking spaces, quality of mental care, or ease of booking an appointment.

A map of topic correlations (Figure 1) is a convenient way to summarize topic modeling results.[4] The topic map allows researchers to make comparisons between topics that have been calculated based on the similarity of words between pairs of topics. The greater the distance and the thinner the connecting line between two topics, the less they tend to occur together in individual reviews. Clusters of related topics are represented by node colors. In this case, red topics relate to themes representing negative experiences, green topics cluster themes associated mostly with positive experiences, and orange topics group themes of patients sharing their personal experience of using specific health services without a strong positive or negative judgment. Topic clusters have been calculated with a community detection algorithm that optimizes clusters to maximize the strength of within-cluster connections relative to between-cluster connections (Blondel et al. 2008). Furthermore, topic node sizes on the map correspond to the prevalence of each topic across the GP reviews.

---

[4] Topic map has been generated with Gephi, a software package for network modelling. For further information about Gephi, please visit: http://gephi.org, viewed on 17 September 2017

| Topic 1 | Topic 2 | Topic 3 |
|---|---|---|
| Distressing phone booking | Good doctors | Bad facilities |
| Topic 4 | Topic 5 | Topic 6 |
| Incompetent | Great vaccine help | Satisfied |
| Topic 7 | Topic 8 | Topic 9 |
| Can't choose doctor | Bad opinions | Appointment impossible |
| Topic 10 | Topic 11 | Topic 12 |
| Decent practice | Hard appointments | Saying thanks |
| Topic 13 | Topic 14 | Topic 15 |
| Poor mental care | Booking roulette | Improved |
| Topic 16 | Topic 17 | Topic 18 |
| Give dignity | Don't listen | Long-term experience |
| Topic 19 | Topic 20 | Topic 21 |
| Male relative went | Paperwork issue | Big changes |
| Topic 22 | Topic 23 | Topic 24 |
| Disappointing | Son treated | [no meaning] |
| Topic 25 | Topic 26 | Topic 27 |
| Great GP | Respectful | Distressing |
| Topic 28 | Topic 29 | Topic 30 |
| Effective help | No appointment | Long-term condition |
| Topic 31 | Topic 32 | Topic 33 |
| Blood test | Long wait times | Long wait |
| Topic 34 | Topic 35 | Topic 36 |
| [meaning not certain] | Hospital referral | Parking problem |
| Topic 37 | Topic 38 | Topic 39 |
| Not helpful | Friendly | Professional |
| Topic 40 | Topic 41 | Topic 42 |
| Repeat prescription | The best ever | Impossible appointment |
| Topic 43 | Topic 44 | Topic 45 |
| Recommend | Long-term happy | Time delay |
| Topic 46 | Topic 47 | Topic 48 |
| Poor experience | Walk-in help | Diabetes check |
| Topic 49 | Topic 50 | Topic 51 |
| Impressive | Out of hours care | Pros and cons |
| Topic 52 | Topic 53 | Topic 54 |
| Empathy | Demand pressure | Upset! |
| Topic 55 | Topic 56 | Topic 57 |
| Really recommend | Lack common sense | [meaning not certain] |

**Table 1: Topic labels** Estimates from 57-topic LDA model, with labeling by the authors.

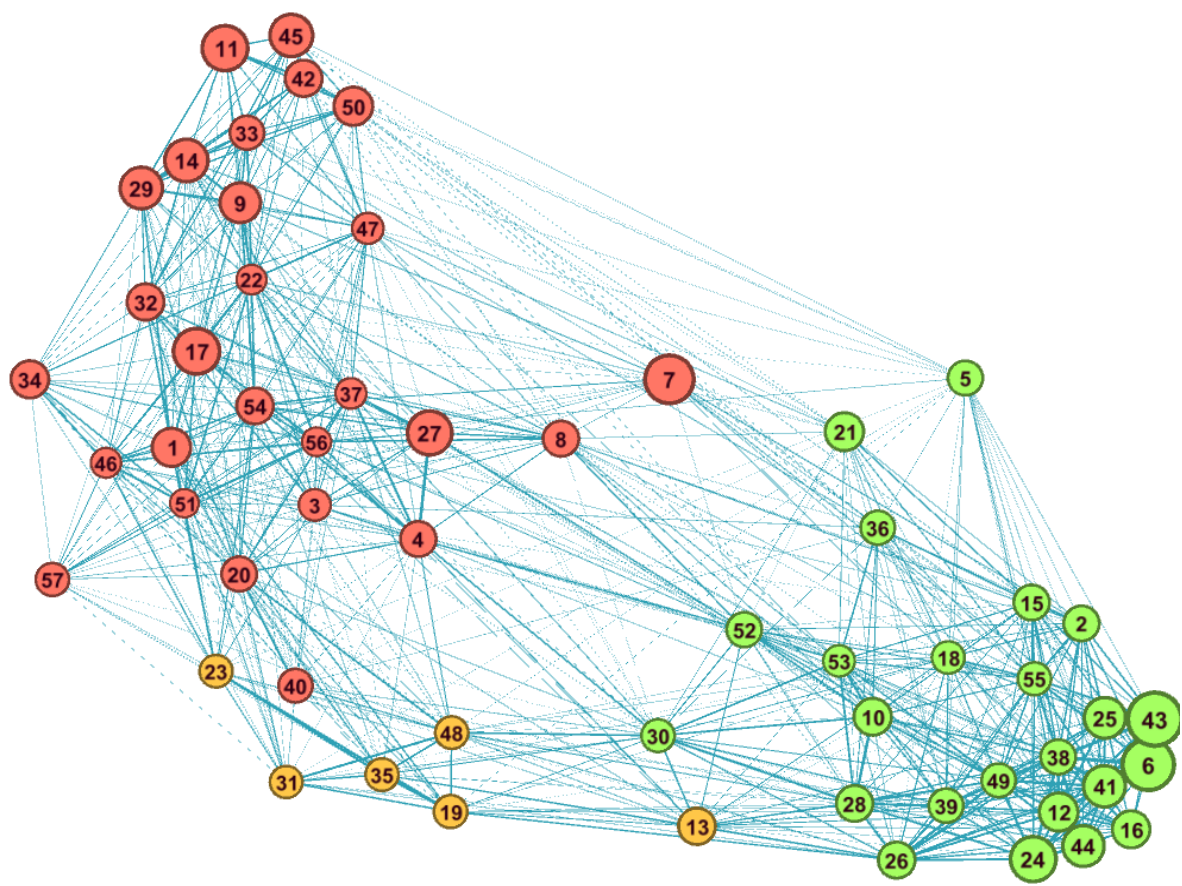

**Figure 1: Topic map for LDA topic model with 57 topics**
*Notes: (1) Topic map illustrates, on a 2-dimensional plane, how similar to one another are 57 topics generated with an LDA topic model from NHS GP practice reviews. Distances between topics are proportional to the differences of the words they contain. The most similar topics in terms of the words they contain tend to be close to one another. (2) Nodes represent individual topics. The bigger the node, the more prevalent is the given topic within the data. (3) The stronger the line connecting a pair of topics, the greater the similarity between the two topics. (4) Node colors indicate clusters to which topics have been assigned. The green cluster contains topics related to positive evaluation of GP service quality. The red cluster groups negative evaluations of GP service quality. The orange cluster groups themes related to specific GP services and personal situations, such as hospital referrals, blood tests, and the mentions of relatives using a given GP practice.*

Figure 1 maps topics with positive GP service evaluations in the bottom-right of the map, having the least in common with topics containing negative evaluations of GP services at the top-left of the map. The second greatest difference is between topics from reviews in which authors focus on their personal service experience in the bottom-left of the map and users who tend to narrate in third person about GP service quality in the top-right of the map. The most common topics include thanks for doctors attending to patients and complaints about the difficulty/impossibility of booking appointments.

## Explaining User Satisfaction with Feedback

As discussed above, the data from GP reviews also contains more standard measures of user satisfaction in the form of Likert scale survey questions. We utilize this feature here to relate our estimates of user satisfaction from topic modeling with more traditional survey-based measures. First, we estimate Random Forest (RF) models where the proportional presence of topic reviews are independent variables, and six Likert-scale ratings are treated as dependent factor variables.

Random Forest is a machine learning algorithm based on building decision trees on bootstrapped (randomly sub-sampled) data with a smaller subset of randomly sampled predictors at each decision node. A large number of trees is grown until a stopping rule is achieved (e.g. minimum of 5 observations in the terminal nodes) and then aggregated for final prediction. Random Forest takes advantage of both weak and strong classifiers, where weak are those that bring a prediction that is slightly better, or just the same as, a random guess. It is valuable due to its interpretation simplicity compared to other machine learning algorithms, and can be used for regression and classification type problems, as well as to model non-linear relationships. For further details on Random Forest models see, for example, Hastie, Tibshirani and Friedman (2001, 587–603). One benefit of using RF models here is that by design they deal with multicollinearity and allow for an unambiguous identification of the relative importance of topics identified with the LDA analysis.

Our multiclass RF model predicts the outcome variables with accuracy ranging from 0.49 on "phone access ease" to 0.77 on "likely to recommend" dimensions. Precision and recall measures vary across "star" levels and dimensions, with the F1-score ranging between close to zero to 0.85. The supplementary materials provide complete set of model quality estimates, and underlying confusion matrices. This variation is partly driven by difference in sample sizes across different models as can be seen from the confusion matrices. Overall, we are capturing some of the relationship between unstructured data (reviews summarized with topic models) and structured data (Likert-scale "star" ratings).

Figure 2 presents the results of the Random Forest model in terms of the importance ranking of independent variables (57 topics) for predicting each individual Likert-scale outcome variable.[5]

Random Forest outcomes (Figure 2) indicate that topics generated from online reviews are related to Likert-scale responses provided by service users, and that satisfaction from multiple aspects of the GP service is, most importantly, related to similar themes present in the reviews. The evaluations of the services on different dimensions are interlinked. It suggests that user satisfaction can be improved among multiple dimensions by adopting a single approach of addressing important, common problems and enhancing the key positive experiences. Emotional positive experiences were the strongest predictor of satisfaction with topic41 ("best ever") as the most important.

The five most common words in this topic (see supplementary materials for details) are: professional, team, efficient, kind, nurse, and thorough. The topic content indicates that the behavior of, and conversations with, staff have the highest influence on how patients evaluate GP services. Positive emotional experience, represented by topics such as "respectful", "impressive", "empathy", "long term happy" and "give dignity", has a relatively weaker impact on Likert-scale ratings, but nonetheless is relatively more important than most topics. On the other hand, the most common problems that affect Likert-scale satisfaction evaluations are service user experiences of mistreatment and perceptions of staff incompetence, followed by the "paperwork issue" topic. Difficulties with making a GP appointment have a moderate importance for predicting the star ratings among negatively charged topics. Topic 14 "booking roulette" has the highest importance for star ratings given on the ease of phone access to the GP practice. A comparison between the topic map (Figure 1) and the Random Forest model (Figure 2) outcomes indicates that patients tend to write reviews more about their overall experience of the GP service over time, including a greater emphasis on communication with the GP practice prior to an appointment. Apart from that, topic 3 "bad facilities" appears consistently as having an impact on survey-based ratings.

---

[5] We show only the top 30 most important predictors to simplify the presentation in the plots.

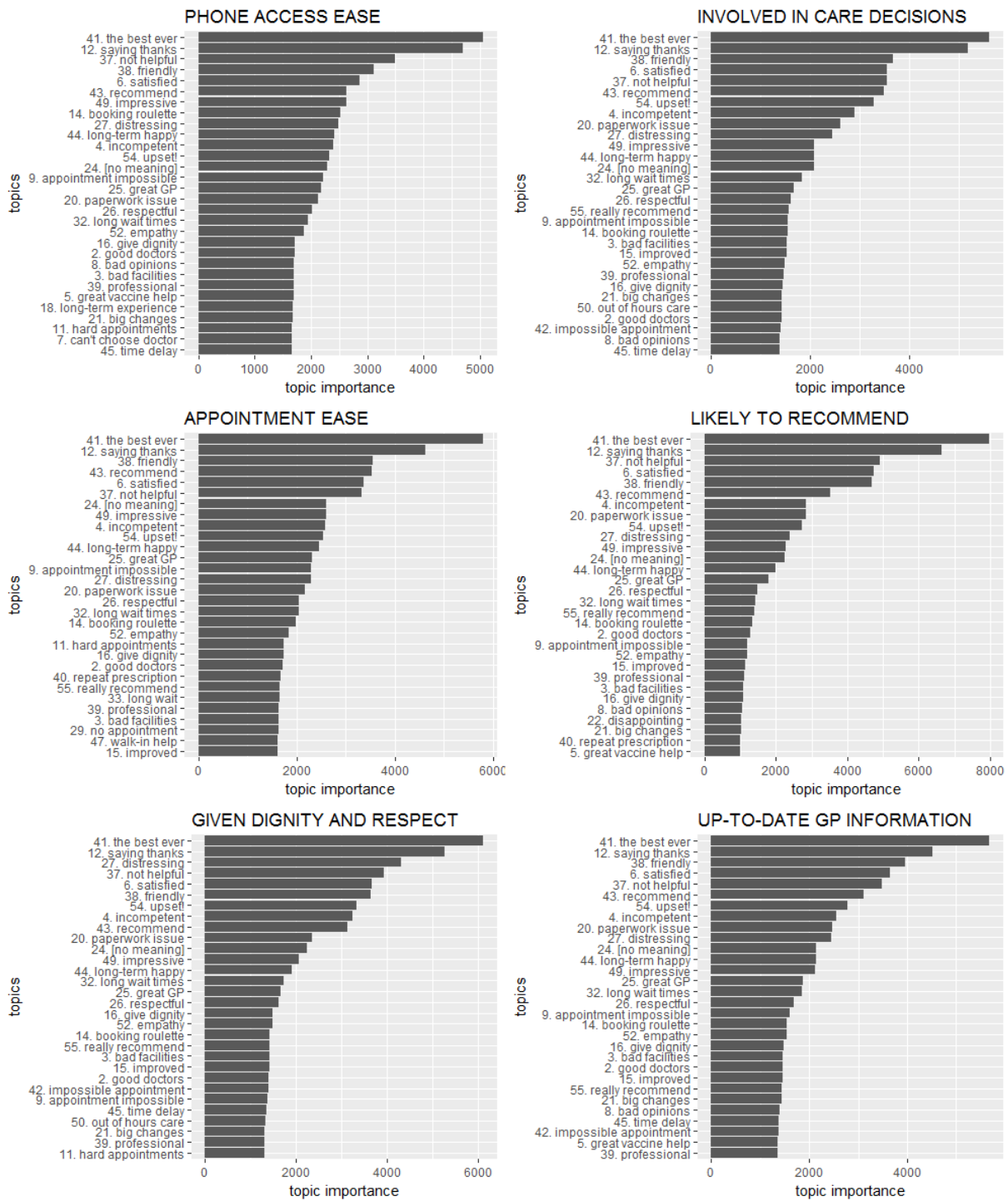

**Figure 2: Random forest model results - importance ranking for topics on six dimensions of GP service quality**

*Notes: (1) Random Forest model outcomes illustrate with horizontal bars the importance of topics (independent variables) for correct prediction of star ratings (dependent variables) given in responses to the 6 Likert-scale survey statements. Star ratings are treated as categorical data. (2) Topic importance represents the average improvement in classification at the moment when a topic is used in the Random Forest model as an independent variable. Model improvement is measured with residual sum of squares. (3) Each sub-figure includes the most important 30 topics for predicting the dependent variable. The omitted 27 topics had scores similar to the included least important topics.*

Overall, Likert-scale evaluations appear firmly related to topics that cluster comments about the administrative and medical service experience from reviews, with ease of communication and the quality of GP facilities also playing a role. More general opinions appear to have little effect on the Likert-scale ratings given the dataset. For example, topics with relatively low importance for predicting the ratings include "lack of common sense" (topic 56), comparisons between GP practices (topic 51), and general observations about NHS services (topic 53).

Topic model outcomes indicate that the relationships of service users with GP staff, as well as the care and respect of GP staff for the users, are the most important for improving satisfaction from GP services. Difficulties with scheduling an appointment, bureaucracy related to GP services, or material shortcomings of GP practices such as outdated facilities are relatively less important. Nonetheless, if GP staff and patients spent less time on administrative efforts and practices had up-to-date facilities and equipment, patient satisfaction would likely also improve. The unique advantage of the findings is that they have been obtained directly from the data without making any assumption as to what makes users satisfied. Insights from analyses like this one could inform public management that helps build worker satisfaction through better relations with service users and improves legitimacy as the provider of high quality public services.

Robustness Analysis

Fixed-effects models were used to establish if, after correcting for variance related to other relevant variables, the statistically significant correlation between topic proportions and star ratings still holds. First, in order to generate panel data that includes control variables, several datasets were merged. Counts of patients registered from each area of England (LSOA, Lower Layer Super Output Area – about 300 households per area) in GP practices in England from 2015[6] were merged with data on levels of deprivation at each LSOA[7] in order to calculate a weighted average of deprivation of patients coming

---

[6] Source: https://data.gov.uk/dataset/numbers-of-patients-registered-at-a-gp-practice-lsoa-level, last visited on 1st August 2017
[7] Source: https://www.gov.uk/government/statistics/english-indices-of-deprivation-2015, last visited on 1st August 2017

to each GP practice, as well as the number of registered patients in each practice. Furthermore, administrative data available for GP practices from 2015 was added to establish which Clinical Commissioning Group (CCG, a mid-level unit of NHS administration) manages disbursement funding for which GP practices[8]. The differences in management style between CCG managers, and also higher-level administrators at regional level, may contribute to changing the circumstances in which patients review their experience at GP practices. The three datasets were combined with topics generated with LDA topic model for each review. The dataset already included Likert-scale rating values for GP service experience and information on when each review was posted. The 5-point Likert-scale star ratings were responses to survey statements: 1) "Are you able to get through to the surgery by telephone?", 2) "Are you able to get an appointment when you want one?", 3) "Do the staff treat you with dignity and respect?", 4) "Does the surgery involve you in decisions about your care and treatment?", 5) "How likely are you to recommend this GP surgery to friends and family if they needed similar care or treatment?", and 6) "This GP practice provides accurate and up to date information on services and opening hours". Dataset mergers resulted in inclusion of 144,192 reviews, with 2,196 reviews originally used to generate 57 topics with a topic model removed from the dataset due to missing attributes. Reviews of new, closed down, and/or less popular GP practices were more likely to be removed from the dataset.

The reviews for which all data was available were transformed into a panel dataset wherein each data point belonged to a different combination of CCG ID and month when a review was posted. Star ratings, topic proportions, deprivation scores, and patient register sizes were averaged for each data point. Panel data were used for calculation of fixed-effects models which accounted for effects of CCG to which a commented-upon GP belonged and the month in which reviews were posted. Likert-scale ratings were used as dependent variables, and topic proportions derived from the content of written reviews as independent variables. Average levels of deprivation among registered patients and GP register sizes were used as control variables. Furthermore, for simplicity, the topics have been grouped into negative and positive clusters, in line with the color coding scheme from Figure 1 in the main paper.

---

[8] Source: http://content.digital.nhs.uk/catalogue/PUB18468, last visited on 1st August 2017

A cluster of neutral topics was also created but has not been used in the fixed-effects models to avoid a multicollinearity problem (topics proportions in reviews always sum to 1). The results of the linear two-way fixed effect (CCG and month) model are presented in Table 2.[9]

|  | Model 1 | Model 2 | Model 3 | Model 4 | Model 5 | Model 6 |
|---|---|---|---|---|---|---|
| **Positive topics** | 1.91 *** | 2.83 *** | 3.70 *** | 4.83 *** | 5.22 *** | 3.33 *** |
|  | (0.26) | (0.24) | (0.26) | (0.26) | (0.30) | (0.24) |
| **Negative topics** | -2.54 *** | -2.78 *** | -1.14 *** | 0.27 | -1.44 *** | -0.99 *** |
|  | (0.26) | (0.24) | (0.27) | (0.27) | (0.31) | (0.25) |
| **Average IMD score** | 0.03 | 0.03 * | 0.02 | 0.04 ** | 0.03 | 0.05 *** |
|  | (0.01) | (0.01) | (0.01) | (0.02) | (0.02) | (0.01) |
| **Number of patients** | -0.00 *** | -0.00 *** | 0.00 | 0.00 | -0.00 | 0.00 |
|  | (0.00) | (0.00) | (0.00) | (0.00) | (0.00) | (0.00) |
| **CCG FE** | YES | YES | YES | YES | YES | YES |
| **Month FE** | YES | YES | YES | YES | YES | YES |
| **R2** | 0.42 | 0.51 | 0.42 | 0.38 | 0.56 | 0.38 |
| **Adj R2** | 0.41 | 0.49 | 0.41 | 0.37 | 0.54 | 0.37 |
| **Num. Obs.** | 9306 | 9306 | 9306 | 9306 | 9306 | 9306 |

**Table 2: Two-way fixed-effects models**

*Notes: Outcomes of fixed-effects models take into account variance in the reviews data that results from differences between Clinical Commissioning Groups (NHS units responsible for funding allocations to GP practices) and monthly time periods when the reviews were posted. Likert-scale star ratings are the dependent variables. Topic proportions within documents are the independent variables. Topic proportions have been clustered into positive, negative, and neutral – in line with the color coding scheme available in Figure 1. The neutral cluster can be predicted with the other two clusters and has not been included in the calculations to avoid the multicollinearity problem. The models included two control variables. The average IMD score (a measure of deprivation, 1 is the best and 10 is the worst) of patients using GP services, as well as a count of how many patients are registered at a reviewed GP practice (a proxy value correcting for GP size). Robust standard errors for coefficients are reported in brackets. Significance: \*\*\* p < 0.001, \*\* p < 0.01, \* p < 0.05*

The results of the two-way fixed-effects models suggest that what patients write is significantly correlated to how they rate their experience. The cluster of positive topics predicts higher star ratings, and the cluster of negative topics predicts lower star ratings. The only exception is in the case of Model 4, where only the positive cluster of topics is a statistically significant predictor of the star ratings. The finding may be due to the fact that reviewers value being involved in care decisions, but their criticism of GP service experience tends to relate to other aspects of care than being involved in care decisions. Additional context variables, such as levels of deprivation in areas which GP practices serve and GP practice sizes, do not meaningfully change the relationship between star rating evaluations and topics.

---

[9] All fixed-effects models were calculated with R programming language, using *plm* package.

## Limitations and Future Work

The study's limitations relate to relative low response rate from the users of GP services. For example, the dataset examined in this study reveals that GP practices received fewer than 20 reviews on average over a period of three and a half years. This makes comparison between individual GP practices infeasible. Instead, we have to limit comparison to mid-level administrative areas. Moreover, the biases in the sample of patient experiences analyzed with the topic model are unknown and hard to predict (e.g., Xiang et al. 2017).

In addition, the data summarization method deployed with topic model has a few known weaknesses. These include: 1) possible misalignment between topic proportional presence in reviews and topic importance for users, 2) unavoidable uncertainty over how many topics to generate to best represent reviews (Grimmer and Stewart 2013; Wallach, Mimno, and Mccallum 2009), as well as 3) crude assumptions made about natural language in the design of the topic model (Grimmer and Stewart 2013; Moody 2016; Winkler et al. 2016).

Therefore, it is advisable to compare topic model results obtained from online reviews with a representative and systematic survey of service user opinions about their service experience. The comparison could help establish the representativeness of topic modeling outcomes. In the instance of the National Health Service in England, the GP Patient Survey is at present the most systematic and regularly collected opinion survey about GP services in England (Cowling, Harris, and Majeed 2015). It could be used every so often to validate topic model outcomes, possibly allowing a decrease in the frequency and cost of data collection for the survey.

In future work, we plan to extend the analysis to the wider NHS system in England and Wales, focusing on hospital reviews. We also intend to explore the underlying assumptions of natural language processing as applied to customer reviews, and assess the sentiment and linguistic properties of individual user feedback.

# Conclusion

Our analysis suggests that topic models are useful for summarizing large numbers of written reviews to identify and analyze the determinants of user satisfaction in public services. The topic modeling outcomes can be similarly complex to conclusions from qualitative studies of similar datasets (e.g., Lopez et al. 2012), but at the same time can be obtained relatively quickly from much larger datasets. Topic models constructed from online reviews could also be helpful in guiding change in public institutions at national and regional level, as opposed to simply informing frontline professionals and service users who read individual reviews. For instance, topic model outcomes like that carried out in this study could help administrators in the National Health Service identify and learn from successful GP practices across England. Patient feedback can be clustered according to the NHS institutions to which it relates, giving insight into patterns of satisfaction and GP management styles across the country. Reviews themselves can also be clustered according to their topical structure and Likert-scale satisfaction levels to understand the prevalent narratives of users about their service experience. For example, some less common narratives may be indicative of distressed users in poor mental health and it would hence be important to better understand where and why this occurs. Other uses of topic models can include analyses of key challenges facing public institutions such as the NHS which could be overcome nationally for all patients. In this sense, the results of this study suggest that many patients express frustration with the difficulty in making GP appointments. Modeling outcomes indicate that a nation-wide online booking system for patients to transparently manage GP appointments may help. In addition, the NHS may also choose to use topic modeling results to generate near real-time insights into patient satisfaction, assessing how decisions affect patients over time, whether some areas suffer from significant shifts in perceived GP service quality, and how the impact of NHS decisions varies in different locations. Finally, topic models can help inform public preferences regarding NHS services. In this way, the public could obtain information about current NHS challenges through the lens of actual GP reviews, as opposed to a limited range of hard figures prepared by the public service provider.

In summary, researchers and public managers can benefit from the introduction of more machine learning algorithms to support inquiries into the determinants of user satisfaction from public services at national and regional levels. Topic model machine learning algorithms can be used to process very large numbers of reviews to generate complex, but at the same time also easy to understand and actionable, insights. Online reviews processed with machine learning can offer a near real-time, dynamically adapting and low cost system for generating user insights. For example, in the instance of public healthcare in England, topic model outcomes obtained from online reviews point to the fact that patients tend to comment proficiently about their difficulties in accessing GP services, but this is not the most important predictor of poor experiences with the health services. Instead, how GP staff treat patients is what determines if users rate their experience highly or not, followed by user experience with paperwork issues related to their treatment. Potentially, a change in communication style by NHS staff (aided by a more convenient online booking service, as suggested above) and streamlined treatment-related formalities could help lift patient satisfaction despite difficulties in getting a GP appointment. Furthermore, thanks to the open source nature of online reviews, topic model outcomes can be made public, in this way responding to the demand for more inclusive decisions about public service provision (O'Leary 2016).

Application of Natural Language Processing to Determine User Satisfaction in Public Services: Online Supplementary Material

# Selecting the number of topics for LDA analysis

Topic models containing from 3 up to 100 topics were calculated from pre-processed data and compared to identify the optimal number of topics for modeling. Following Roberts et al. (2015), 97 topic models were evaluated with semantic coherence (the rate at which topic's most common words tend to occur together in the same reviews) and exclusivity (the rate at which most common terms are exclusive to individual topics) scores. The model with 57 topics had the best combination of semantic coherence and exclusivity scores out of all models. It had the highest semantic coherence score and one of the highest exclusivity scores (see Figure A1) which means the model with 57 topics has generated the most distinct and semantically coherent set of key themes.

**Figure A1: Semantic coherence and exclusivity scores for calculated topic models**

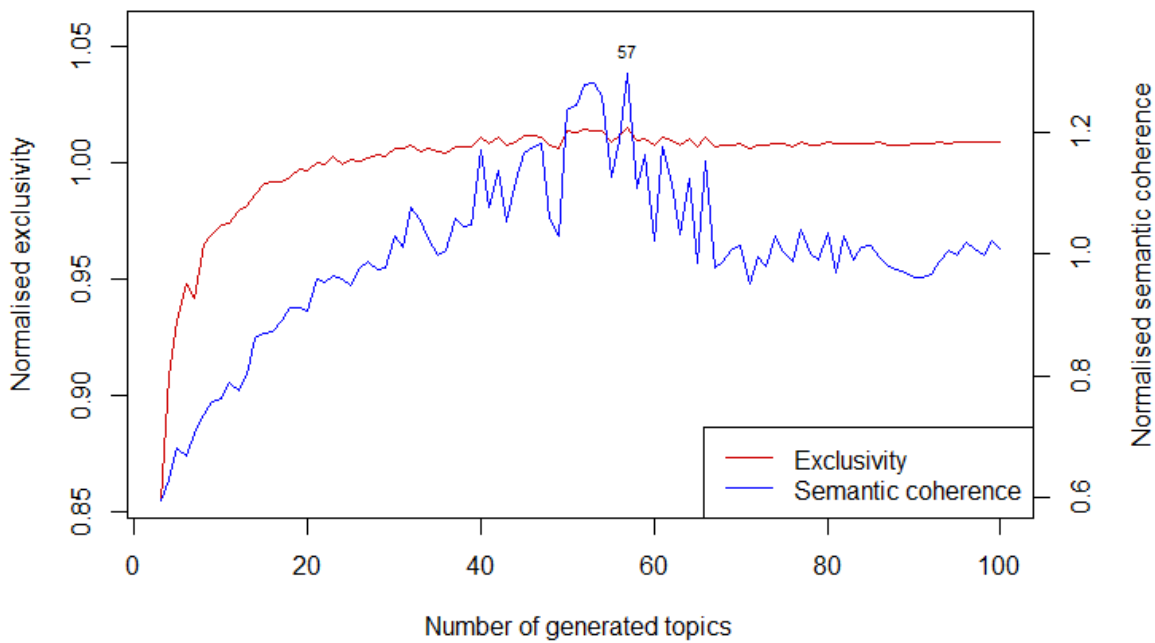

*Notes: (1) The illustration portrays semantic coherence (the rate at which each topic's most common words tend to occur together in the same reviews) and exclusivity (the rate at which most common terms are exclusive to individual topics) for topic models with up to 100 generated topics. Higher semantic coherence and exclusivity scores tend to correlate with higher perceived quality of generated topics. (2) Scores were normalized by dividing individual model scores by average scores for all models*

# Explanation of topic labelling

The 57 topics generated with the chosen STM topic model have been labelled according to the most frequently occurring words in topics as well as the written reviews which are representative of each topic. Table A1 below lists 7 most frequently occurring terms for each topic and Table A2 includes labels assigned to each topic together with a review representing the topic. Representative reviews have been identified by high proportion of terms within reviews classified into a given topic. Each topic not meaningfully related to GP service experience based on inspection of the words and prominent reviews representing it has been highlighted in yellow.

**Table A1: Most prominent words for STM model with 57 topics**

Topic 1 Top Words:
    call, back, got, today, told, rang, daughter
Topic 2 Top Words:
    good, doctor, keep, surgeri, better, problem, realli
Topic 3 Top Words:
    room, hear, door, look, stop, stand, one
Topic 4 Top Words:
    complet, lack, seem, avoid, total, simpli, dismiss
Topic 5 Top Words:
    nurs, clinic, surgeri, children, need, also, time
Topic 6 Top Words:
    staff, alway, help, recept, found, polit, pleasant
Topic 7 Top Words:
    doctor, see, say, never, want, one, problem
Topic 8 Top Words:
    peopl, review, think, one, seem, surgeri, read
Topic 9 Top Words:
    get, appoint, ring, never, tri, gone, will
Topic 10 Top Words:
    patient, practic, general, mani, staff, admin, demand
Topic 11 Top Words:
    appoint, book, get, week, emerg, urgent, day
Topic 12 Top Words:
    thank, support, care, help, famili, grate, appreci
Topic 13 Top Words:
    health, issu, ill, life, feel, condit, serious
Topic 14 Top Words:
    phone, call, answer, tri, get, line, telephon
Topic 15 Top Words:
    surgeri, servic, good, new, improv, pleas, hous
Topic 16 Top Words:
    treat, respect, way, surgeri, mother, patient, like
Topic 17 Top Words:
    apo, amp, don, can, get, doesn, isn
Topic 18 Top Words:
    doctor, surgeri, problem, time, reason, thought, also
Topic 19 Top Words:
    hospit, home, visit, doctor, arrang, husband, immedi
Topic 20 Top Words:
    ask, told, said, letter, form, regist, went
Topic 21 Top Words:
    year, chang, now, surgeri, last, sinc, differ
Topic 22 Top Words:

        can, get, amp, just, apo, time, actual  
Topic 23 Top Words:  
        pain, son, went, infect, doctor, antibiot, prescrib  
Topic 24 Top Words:  
        care, servic, excel, receiv, provid, treatment, high  
Topic 25 Top Words:  
        feel, much, doctor, noth, made, explain, felt  
Topic 26 Top Words:  
        treatment, advic, concern, doctor, quick, wife, recent  
Topic 27 Top Words:  
        rude, staff, recept, receptionist, unhelp, attitud, train  
Topic 28 Top Words:  
        practic, consult, requir, patient, continu, telephon, offer  
Topic 29 Top Words:  
        appoint, day, told, morn, book, next, week  
Topic 30 Top Words:  
        despit, term, long, medic, suggest, regular, becom  
Topic 31 Top Words:  
        test, blood, result, done, doctor, take, taken  
Topic 32 Top Words:  
        wait, minut, receptionist, ask, arriv, anoth, min  
Topic 33 Top Words:  
        appoint, get, time, see, need, can, week  
Topic 34 Top Words:  
        amp, apo, couldn, just, now, yet, get  
Topic 35 Top Words:  
        refer, referr, hospit, diagnos, symptom, specialist, month  
Topic 36 Top Words:  
        area, surgeri, use, new, park, move, choic  
Topic 37 Top Words:  
        manag, pay, complaint, nhs, amp, privat, will  
Topic 38 Top Words:  
        practic, gps, staff, approach, good, except, alway  
Topic 39 Top Words:  
        medic, centr, given, short, doctor, occas, surgeri  
Topic 40 Top Words:  
        prescript, repeat, request, medic, order, pharmaci, surgeri  
Topic 41 Top Words:  
        profession, team, effici, kind, nurs, enough, thorough  
Topic 42 Top Words:  
        appoint, get, day, imposs, see, abl, make  
Topic 43 Top Words:  
        alway, friend, recommend, help, surgeri, staff, happi  
Topic 44 Top Words:  
        practic, year, move, regist, care, famili, medic  
Topic 45 Top Words:  
        time, wait, appoint, hour, seen, run, late  
Topic 46 Top Words:  
        apo, amp, didn, wasn, wouldn, doctor, even  
Topic 47 Top Words:  
        surgeri, walk, need, take, will, get, can  
Topic 48 Top Words:  
        inform, check, record, procedur, diabet, date, advis  
Topic 49 Top Words:  
        care, import, posit, confid, feel, practis, person  
Topic 50 Top Words:  
        work, appoint, system, open, get, can, time  
Topic 51 Top Words:  
        apo, amp, doctor, pretti, one, time, thing  
Topic 52 Top Words:

```
                particular, however, one, often, littl, seem, rather
Topic 53 Top Words:
                nhs, surgeri, patient, part, set, countri, time
Topic 54 Top Words:
                just, even, wrong, absolut, wors, bad, doctor
Topic 55 Top Words:
                surgeri, doctor, need, time, find, patient, yes
Topic 56 Top Words:
                apo, amp, know, can, just, like, exact
Topic 57 Top Words:
                amp, quot, apo, didn, ask, said, say
```

**Table A2: Topic labels with representative reviews**

| Topic 1 | Topic 2 | Topic 3 |
|---|---|---|
| Distressing phone booking | Good doctors | Bad facilities |
| "i called them yesterday for a appointment and was told a doctor would call me back 30 hours later still no call""i called them yesterday for a appointment and was told a doctor would call me back 30 hours later still no call" | "Very good surgery. Doctors are nice and understanding, and they listen to your problems. Very pleased so far." | "No privacy at reception. No access to chilled drinking water. Dated chairs/ waiting room. Hand gel? Stuffy warm atmosphere. No ventilation. Good staff but facilities really needs bringing up to date." |
| Topic 4 | Topic 5 | Topic 6 |
| Incompetent | Great vaccine help | Satisfied |
| "Left without important medication due to incompetent staff and unhelpful doctor. Totally disorganised surgery and a disgrace to the NHS. Avoid." | "Well organised winter flu vaccinations in 2014. I was in and out of the surgery within 20 minutes." | "I have allways found the doctors, nurses and reception staff helpfull and pleasant." |
| Topic 7 | Topic 8 | Topic 9 |
| Can't choose doctor | Bad opinions | Appointment impossible |
| "can not get in to see any doctor most of the time. can never see my own doctor.if you get in you only see a learning doctor." | "It surprises me when I read the other reviews as I have always been treated well. My only negative is sometimes the phones are very busy and it can take a while to get through." | "you can never get an appointment, when u ring up at 8.30am u cant get threw and when u finally get threw they never have any appointments ......" |
| Topic 10 | Topic 11 | Topic 12 |
| Decent practice | Hard appointments | Saying thanks |
| "The Practice is very good indeed and generally meets the needs of patients." | "It is very difficult to get an appointment with my doctor unle | "Excellent care from the doctor by going extra mile to help me    Many thanks for your care" |
| | "ss I ring at least two weeks in advance." | |
| Topic 13 | Topic 14 | Topic 15 |
| Poor mental care | Booking roulette | Improved |
| "In a few words:- I felt sharply put down, such that neither I - nor my ongoing complex health issues, would be helpfully reviewed." | "If u start ringing 8am line is busy after 10 minutes when u lucky and somebody answer no appointment today. ..." | "I switched to awburn house approx 18 months ago. What a breath of fresh air. Very professional and accessible. The doctor is great." |
| Topic 16 | Topic 17 | Topic 18 |
| Give dignity | Don't listen | Long-term experience |
| "Everyone at the surgery from receptionist to the doctors, age no exception everyone treated with great respect and dignity" | "The GP at this surgery doesn&apos;t ask how you are, don&apos;t care how you are and always tells you that nothing is wrong with you, won&apos;t help you and doesn&apos;t know who you are each time you go. Useless." | "I have bean looked after at this surgery for over 10 years by only two doctors. First by one doctor up to this year. I consider my self being very lucky to have had such a caring doctor. Now I am again lucky to have another doctor whom also make me feel that they also care about their patients. I hope that all the doctors will eventually settle and look after all their patients." |
| Topic 19 | Topic 20 | Topic 21 |
| Male relative went | Paperwork issue | Big changes |
| "Tried to ring the doctors from 8.00 to 8.30 because my husband was unwell, could not get through at all so he ended up visiting another practice who wondered if he had suffered a mini stroke and referred him to the hospital." | "was forced to take diazipapane that i no longer want to take due to addiction issues was told by GP to take them anyway and when i asked for a bus pass form to be signed and stamped was told it would take 3 weeks! to complete a box on a form" | "when I joined this surgery a 5 years ago I could see a regular doctor but for the last few years all you get is locum who are just going through the motions I have requested a couple of times to speak to the management of the practice but not been sorted yet" |
| Topic 22 | Topic 23 | Topic 24 |
| Disappointing | Son treated | [no meaning] |
| "My Wife had an appointment today just to be told none of the doctors had turned up, very disappointed, she eventually got turned away, my 6 month old Daughter has an appointment this afternoon, they don&apos;t know if the doctors are going to turn up so probably won&apos;t happen, this surgery is getting worse, h | "I was told that I had Planter Fasciitis in my heel and a trapped nerve causing the pain in my back and arm. I was given treatment for both of these issues and am now free from pain in both." | "isPermaLink=\"false\">492722</guid><link>http://www.nhs.uk/Services/GP/ReviewsAndRatings/DefaultView.aspx?id=37275</link><a10:author><a10:name>Anonymous</a10:name></a10:author><category domain=\"commentType\">comment</category><title>Excellent services provided.</title>... < truncated> |
| ow can they get away with just not turning up." | | |
| Topic 25 | Topic 26 | Topic 27 |
| Great GP | Respectful | Distressing |
| "I saw the doctor and was really impressed. They listened to me, made sure I understood my treatment options and made me feel very relaxed." | "I have recently visited the surgery and was given appointment very quickly the doctor was very professional and very respectful and had we had a very good discuss" | "Patronising, unhelpful and mostly incompetent. Worst reception staff ever and unprofessional, need training in confidentiality." |
| Topic 28 | Topic 29 | Topic 30 |
| Effective help | No appoinment | Long-term condition |
| "A super local surgery now ran via Formby surgery offering a balance of essential local needs to an increasingly vulnerable community &amp;amp; access to more specialist treatment than hospitals can offer" | "Call to make an appointment and told there are none available for over 2 weeks. Ask to make one and then told book not open, call back in a week to make an appointment." | "Staff becoming increasingly rude lack of understanding especially long term conditions thought they was a good choice but definitely been proved wrong!" |
| Topic 31 | Topic 32 | Topic 33 |
| Blood test | Long wait times | Long wait |
| "I normally have huge bruises after having a blood sample taken. This month, the lovely nurse practitioner has taken two blood samples from me with no bruising/marks whatsoever." | "I was told I would have a 30 minute wait, however this turned into a 2-30hour wait whilst people came and got seen before me. When I asked how much longer I was rudely told that I would be the last of the day. Which make me ask why would they tell me it was a 30 minute wait. From this I got a rude and blunt reply." | "Every time I phone up for a appointment it take two to four weeks to get one the last time it took over a month to get one. What`s happened to 24 hours it's a joke too many patients ??" |
| Topic 34 | Topic 35 | Topic 36 |
| [meaning not certain] | Hospital referral | Parking problem |
| "Very happy with this gp. Have been to a few all have let me down. this hasn&apos;t as of yet." | "My wife was informed she had enlarged ovary. The GP referred for hospital. At hospital the consultant dismissed the diagnosis. After 2 weeks of worrying about cancer etc... It was a big relief." | "Having undergone extensive knee surgery I can drive but cant walk distance. Not disabled so cant use disabled parking. No on street parking. Nelson st car park full of police cars. Have to use tudor sq car park . Not happy" |
| Topic 37 | Topic 38 | Topic 39 |

| Not helpful | Friendly | Professional |
|---|---|---|
| "This surgery is possibly the worst I have come across. Reception staff are rude and have no interest in welfare of the patients. It seems like they have been forced in to doing their jobs for which we the tax payers pay. Getting an appointment is a nightmare. The list goes on..." | "I find all contacts with staff, including reception, dispensary, GPs and nursing, to be friendly, sensitive and helpful. It far exceeds my experience at other practices previously. My experience is with the Sudbury surgery in particular." | "Whenever I pay a visit to the Trinity Medical Centre I find that the medical care given by the GP is excellent. And the care given by all of the staff is very professional. This is a first class Health Centre." |
| Topic 40 | Topic 41 | Topic 42 |
| Repeat prescription | The best ever | Impossible appointment |
| "I have a monthly prescription which is requested by my local chemist. On Tuesday a request was sent through to the surgery and once again no Prescription has arrived at the chemist. . Very disappointed" | "The practice nurse is an absolute gem, efficient, calm and approachable, a fantastic asset to the team," | "Impossible to get appointment within 7 days. Sometimes can get after 14days." |
| Topic 43 | Topic 44 | Topic 45 |
| Recommend | Long-term happy | Time delay |
| "Great surgery and really helpful and friendly staff. Would definitely recommend to family and friends" | "My family has been registered with Primrose Hill Surgery for 26 years. We consider ourselves fortunate to have such a caring and helpful local practice." | "Very long waiting times once booked in for appointments. Waits for more than an hour on occasion, especially evenings." |
| Topic 46 | Topic 47 | Topic 48 |
| Poor experience | Walk-in help | Diabetes check |
| "I had an appointment at the end of December 2015. I ended with the doctor telling me to be careful as I could become diabetic, I&apos;ve been diabetic since 1999, they hadn&apos;t even read my file." | "If you are a registered patient and take ill after 8 am in the morning you have to wait until the next morning before you can book an appointment. Although there is a walk-in, registered patients are not allowed to use it. Therefore patients not from this surgery have better access to seeing a doctor. The only option is to go to another walk in. A ludicrous situation." | "Diabetic care is excellent with the 6 monthly checks being full and comprehensive. Practice nurse is informed and encouraging meaning that I am still on only diet and exercise after 5 years with blood results being fully explained." |
| Topic 49 | Topic 50 | Topic 51 |
| Impressive | Out of hours care | Pros and cons |

| "I am truly impressed with the staff and service and truly personal touch I receive. I am made to feel like an individual as opposed to a number.\nThank you" | "I have been trying to get an appointment for over 2 weeks and although the surgery has extended opening hours on a Monday evening it is impossible to get an appointment outside working hours." | "The Doctor doesn&apos;t know what his talking about sometimes, my daughter had chicken pox confirmed by another hospital, in which the chicken pox was clear, and the doctor said it was eczema, how bad is that. Something needs to be done." |
|---|---|---|
| Topic 52 | Topic 53 | Topic 54 |
| Empathy | Demand pressure | Upset! |
| "This place is mainly run by locums. Generally they lack compassion or even empathy and not really interested in you. Receptionist are rude and uninformed." | "However busy they are, the doctor always makes you feel at ease and not as though you are on a conveyor belt. Clearly a very experienced medical practitioner who understands I prefer a down to earth approach. They are kind but firm and has a great sense of humour . The nurses are clearly part of the team and are always on top of their game both technically and their approach to people   Long may this continue" | "Whenever we call to GP, no one is attending the phone call. Even if we go directly, they are not giving proper response. This is happening everytime. Really we got upset." |
| Topic 55 | Topic 56 | Topic 57 |
| Really recommend | Lack common sense | [meaning not certain] |
| "I have been using this surgery for 3 years now, I find the receptionists to be very helpful (which I find is quite unusual at a doctors surgery) they have always found a way to fit myself and my family in for an appointment! Thank you!" | "For the past few months this surgery has gone downhill we all wanna know what&apos;s happened to our GP they are our doctor instead we get another doctor and the reception staff are not helping at all the only thing we are told is our GP isn&apos;t practising anymore we have the right to know" | "organisation of the system was just great this year and we &amp;quot;flew&amp;quot; through the surgery in just 5 minutes. Please keep this system for all such &amp;quot;mass&amp;quot; patient programs" |

The features extracted from text reviews with LDA topic model relate to a range of experiences of patients. Some relate to whether GP staff were helpful or not, to cases of perceived misdiagnosis and difficulties in having a GP appointment. Several topics also offered assessments of the situation of GP services, or were about comparisons between different staff members or GP practices. Other topics covered evaluations of GP facilities such as toilets and information online about the practice. Finally, several topics 24, 34 and 57 have been generated which relate more to the choices of words used in specific comments than a discernible aspects of GP services. The topics had a varying prevalence across the GP reviews dataset (see Figure A2 below), from less than 1% of all tokens in the dataset to under 4%. The model clustered reviews according the choices of vocabulary used by reviewers. Topic 43 "recommend" has been the most prevalent of all of them, followed by topic 6 "satisfied". Topics about the difficulty of scheduling an appointment (1, 7, 9, 11, 29, 32, 33, 42) also frequently featured in reviews, cumulatively constituting about 17% of all content in reviews on average.

Figure A2 presents proportion of appearance in the corpus for each topic in our estimation.

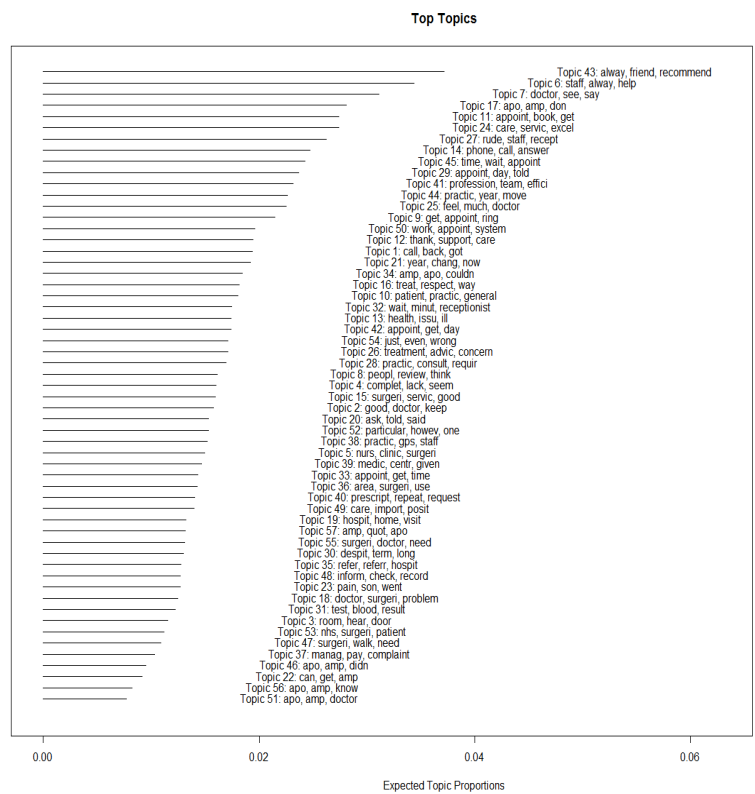

**Figure A2: Topic proportions in the GP reviews dataset**

# Examination of LDA models with 5, 10, 20 and 30 topics

Alternative LDA models with fewer topics (5, 10, 20 and 30), which fall in where the exclusivity rate goes up (Figure 1 from main paper), have been investigated.

**Figure 1 (main paper): Semantic coherence and exclusivity scores for calculated topic models**

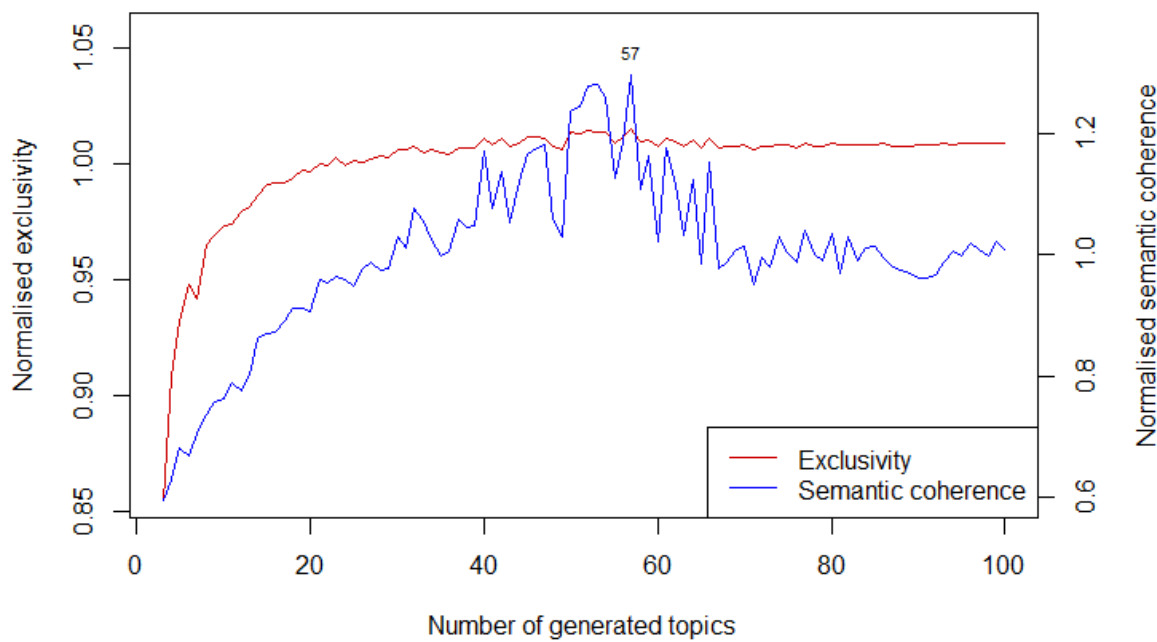

Key findings:

- It appears that the models with fewer topics retain thematic duplicates if some general theme is very common in reviews (see Tables A3-A6 below). Even the LDA model with 5 topics has 2 of them covering the issue of rudeness and staff not listening to the patients (Table A3)
- Simultaneously, themes which may be of interest to NHS decision makers but are more specific to individuals, such as treatment of mental problems, parking access or hospital referrals, relatively quickly disappear from the lists of topics in models. Themes from the 57-topic LDA model 'mental treatment' and 'long-term condition' appear to have portions of their vocabularies combined into one topic in the 30-topic LDA model (Table A6). Model with 20 topics (Table A5) appears to compress both of those subjects together with a 'serious health condition' theme. Similarly, the 30-topic model clusters feedback about blood tests and hospital referrals into one subject, while in a 57-topic model those issues would be kept separate. Comments present in the 57-topic model such as 'bad facilities', 'paperwork issues' or 'can't choose doctor' disappear altogether in models with fewer topics.

- Linear regressions, lasso models and cross-validation calculations have also been carried out for the same set of models as in the draft paper. The results were compared (Tables A7 and A8). Cross-validation errors for linear regressions appear to be the lowest for the topic model with 30 topics but overall a lasso model carried out on 57 topics still has the lowest average prediction error. All regression and lasso models perform better than the baseline, i.e. the standard deviation from average star rating.
- Lasso models with lower topic numbers appear to have very similar but in most cases slightly greater cross-validation errors compared to corresponding linear regression models (Tables A7 and A8)
- It is better to avoid comparisons between topics from different models based on their top words at face value. Topics with seemingly overlapping meanings have very different coefficient values in regression models with the same dependent variables. For example, topics from the 57-topic LDA model 'mental treatment' and 'long-term condition' have negative or no coefficient for predicting in lasso model outcomes (Table A9). In the 30-topic LDA model, topic 21 with an overlapping set of top words had no coefficient in lasso models for every dependent variable (Table A10).

**Conclusions:**

- Some valuable information gets lost when a topic model is calculated with fewer key topics, especially on less talked-about subjects which nonetheless may be important to understanding of service user satisfaction
- There is no single best model with LDA but definitely models with 5 and 10 models have much higher cross-validation errors than the rest. A model with more topics gives insight into more detail but at the same time some popular topics are over-represented and cloud interpretability of model outcomes.

**Table A3: Most prominent words for LDA model with 5 topics**

Topic 1 Top Words:
    appoint, get, time, call, day, wait, phone
Topic 2 Top Words:
    receptionist, doctor, one, patient, like, rude, recept
Topic 3 Top Words:
    doctor, prescript, medic, test, hospit, ask, told
Topic 4 Top Words:
    doctor, surgeri, alway, practic, staff, help, care
Topic 5 Top Words:
    amp, apo, quot, don, doctor, just, didn

**Table A4: Most prominent words for LDA model with 10 topics**

Topic 1 Top Words:
    doctor, hospit, pain, nurs, went, week, saw
Topic 2 Top Words:
    surgeri, staff, recept, doctor, good, problem, year
Topic 3 Top Words:
    appoint, get, day, book, phone, tri, time
Topic 4 Top Words:
    medic, test, blood, result, made, feel, without
Topic 5 Top Words:
    care, servic, medic, health, treatment, receiv, recent
Topic 6 Top Words:
    call, told, receptionist, back, prescript, ask, surgeri
Topic 7 Top Words:
    patient, practic, manag, nhs, review, servic, howev
Topic 8 Top Words:
    alway, help, doctor, care, friend, nurs, practic
Topic 9 Top Words:
    doctor, see, can, time, one, wait, never
Topic 10 Top Words:
    amp, apo, quot, don, didn, like, rude

**Table A5: Most prominent words for LDA model with 20 topics**

Topic 1 Top Words:
    quot, said, didn, nurs, told, son, daughter
Topic 2 Top Words:
    time, wait, never, hour, see, minut, get
Topic 3 Top Words:
    alway, staff, help, surgeri, friend, recept, good
Topic 4 Top Words:
    use, good, difficult, often, howev, telephon, sometim
Topic 5 Top Words:
    appoint, get, day, need, can, work, see
Topic 6 Top Words:
    receptionist, rude, staff, recept, peopl, one, speak
Topic 7 Top Words:
    doctor, like, feel, realli, say, one, see
Topic 8 Top Words:
    prescript, medic, repeat, request, inform, letter, order
Topic 9 Top Words:
    care, thank, excel, profession, receiv, nurs, famili
Topic 10 Top Words:
    patient, manag, poor, room, number, clear, rather
Topic 11 Top Words:
    test, hospit, back, blood, told, result, went
Topic 12 Top Words:
    just, ask, even, want, know, tell, give
Topic 13 Top Words:
    surgeri, will, now, month, anoth, walk, two
Topic 14 Top Words:
    practic, servic, patient, medic, provid, health, centr

Topic 15 Top Words:
    doctor, listen, time, problem, seen, visit, also
Topic 16 Top Words:
    year, surgeri, move, regist, recent, sinc, area
Topic 17 Top Words:
    call, phone, get, tri, appoint, told, book
Topic 18 Top Words:
    experi, review, mani, issu, long, pressur, comment
Topic 19 Top Words:
    amp, apo, don, can, doesn, wouldn, couldn
Topic 20 Top Words:
    health, condit, issu, serious, treatment, consult, sever

**Table A6: Most prominent words for LDA model with 30 topics**

Topic 1 Top Words:
    doctor, pain, went, daughter, saw, son, took
Topic 2 Top Words:
    even, answer, tell, someon, min, phone, just
Topic 3 Top Words:
    staff, recept, good, surgeri, member, keep, busi
Topic 4 Top Words:
    realli, one, better, think, doctor, lot, peopl
Topic 5 Top Words:
    use, surgeri, improv, children, also, open, short
Topic 6 Top Words:
    appoint, need, work, day, abl, offer, difficult
Topic 7 Top Words:
    help, alway, doctor, best, great, surgeri, happi
Topic 8 Top Words:
    get, phone, tri, appoint, can, ring, day
Topic 9 Top Words:
    left, attitud, complet, point, avoid, pay, ignor
Topic 10 Top Words:
    just, like, know, bad, someth, wrong, anyth
Topic 11 Top Words:
    doctor, see, never, will, say, one, want
Topic 12 Top Words:
    test, hospit, blood, result, refer, referr, check
Topic 13 Top Words:
    care, thank, treatment, receiv, support, team, medic
Topic 14 Top Words:
    year, doctor, surgeri, mani, well, seen, quick
Topic 15 Top Words:
    appoint, book, system, avail, make, onlin, day
Topic 16 Top Words:
    patient, person, receptionist, peopl, speak, room, name
Topic 17 Top Words:
    prescript, repeat, request, medic, letter, order, inform
Topic 18 Top Words:
    quot, ask, said, amp, didn, couldn, told
Topic 19 Top Words:
    surgeri, year, now, regist, move, new, chang
Topic 20 Top Words:
    manag, patient, review, practic, poor, nhs, complaint
Topic 21 Top Words:
    medic, health, issu, condit, problem, concern, discuss

Topic 22 Top Words:
    servic, practic, patient, provid, gps, experi, nhs
Topic 23 Top Words:
    alway, friend, recommend, excel, profession, practic, famili
Topic 24 Top Words:
    wait, time, hour, minut, appoint, seen, see
Topic 25 Top Words:
    call, told, back, week, day, next, today
Topic 26 Top Words:
    nurs, visit, time, two, last, clinic, surgeri
Topic 27 Top Words:
    feel, treat, listen, doctor, respect, made, make
Topic 28 Top Words:
    seem, actual, peopl, let, one, shame, can
Topic 29 Top Words:
    receptionist, rude, ever, absolut, unhelp, one, surgeri
Topic 30 Top Words:
    amp, apo, don, doesn, can, wouldn, isn

**Table A7: 5-fold cross-validation errors for linear regression models**

*Notes:*
*- In the illustration below, star ratings are the dependent variables. Topic proportions in documents are the independent variables*
*- The lower is the average prediction error, the better the model. Green indicates the best model*

| # of topics | Model 1 | Model 2 | Model 3 | Model 4 | Model 5 | Model 6 | Mean |
|---|---|---|---|---|---|---|---|
| 5 | 1.230856 | 1.233476 | 1.285702 | 1.446738 | 1.415361 | 1.417955 | 1.338348 |
| 10 | 1.215537 | 1.188884 | 1.320514 | 1.38704 | 1.325138 | 1.365078 | 1.300365 |
| 20 | 1.092372 | **1.11721** | 1.076852 | 1.287285 | 1.200231 | 1.249655 | 1.170601 |
| 30 | 1.085127 | 1.118057 | **1.069932** | **1.265219** | **1.187555** | **1.249054** | **1.162491** |
| 57 | **1.084** | 1.35 | 1.26 | 1.42 | 1.56 | 1.37 | 1.340667 |
| Standard deviations of star ratings | 1.469908 | 1.60779 | 1.583002 | 1.602535 | 1.840799 | 1.540219 | 1.607376 |

**Table A8: 5-fold cross-validation errors for lasso models**

*Notes:*
*- In the illustration below, star ratings are the dependent variables. Topic proportions in documents are the independent variables*
*- The lower is the average prediction error, the better the mode. Green indicates the best model*

| # of topics | Model 1 | Model 2 | Model 3 | Model 4 | Model 5 | Model 6 | Mean |
|---|---|---|---|---|---|---|---|
| 5 | 1.230897 | 1.233508 | 1.285723 | 1.446830 | 1.415512 | 1.417937 | 1.338401 |
| 10 | 1.215648 | 1.189027 | 1.320489 | 1.387091 | 1.325263 | 1.365115 | 1.300438 |
| 20 | 1.092463 | 1.117379 | 1.076942 | 1.287586 | 1.200294 | 1.249768 | 1.170738 |
| 30 | 1.085128 | 1.118171 | 1.070124 | 1.265289 | 1.187758 | 1.249096 | 1.162594 |
| 57 | 1.083 | 1.10 | 1.06 | 1.26 | 1.18 | 1.25 | 1.1555 |
| Standard deviations of star ratings | 1.469908 | 1.60779 | 1.583002 | 1.602535 | 1.840799 | 1.540219 | 1.607376 |

**Table A9: 57-topic LDA – Top predictors for lasso models where star ratings are the dependent variable**

*Notes:*
- *Predictors for each model are ranked by how different are their coefficients from 0. The ranks are provided in brackets.*
- *Magnitudes of topics from 0 correspond to how important is each topic for predicting the dependent variables*
- *Topics without rank and coefficient values are not statistically significant predictors*

| Topics | PHONE ACCESS EASE | | APPOINTMENT EASE | | GIVEN DIGNITY AND RESPECT | | INVOLVED IN CARE DECISIONS | | LIKELY TO RECOMMEND | | UP-TO-DATE GP INFORMATION | |
|---|---|---|---|---|---|---|---|---|---|---|---|---|
| | Model 1 | rank | Model 2 | rank | Model 3 | rank | Model 4 | rank | Model 5 | rank | Model 6 | rank |
| 55. really recommend | 9.42 | 1 | 17.24 | 1 | 12.39 | 2 | 12.82 | 1 | 22.88 | 1 | 11.71 | 1 |
| 37. not helpful | -8.71 | 2 | -8.22 | 2 | -9.98 | 3 | -10.6 | 3 | -10.0 | 3 | -10.7 | 2 |
| 4. incompetent | -6.25 | 4 | -7.50 | 4 | -13.6 | 1 | -12.4 | 2 | -13.9 | 2 | -9.57 | 3 |
| 54. upset! | -6.13 | 5 | -6.13 | 7 | -9.42 | 4 | -10.4 | 4 | -8.98 | 5 | -8.79 | 4 |
| 2. good doctors | 5.11 | 7 | 7.25 | 5 | 6.49 | 6 | 6.61 | 5 | 9.88 | 4 | 6.18 | 5 |
| 26. respectful | 4.34 | 10 | 5.85 | 8 | 4.79 | 8 | 5.93 | 6 | 8.13 | 6 | 4.87 | 9 |
| 47. walk-in help | -3.85 | 11 | -7.09 | 6 | -3.77 | 12 | -4.27 | 10 | -6.61 | 9 | -5.25 | 7 |
| 9. appointment impossible | -5.62 | 6 | -5.12 | 10 | -4.08 | 10 | -4.62 | 8 | -4.65 | 14 | -4.89 | 8 |
| 20. paperwork issue | -2.88 | 15 | -3.58 | 17 | -5.74 | 7 | -5.19 | 7 | -4.94 | 13 | -5.39 | 6 |
| 27. distressing | -4.35 | 9 | -4.15 | 12 | -8.73 | 5 | -3.66 | 14 | -4.40 | 16 | -4.46 | 10 |
| 49. impressive | 3.31 | 14 | 4.14 | 13 | 4.26 | 9 | 4.52 | 9 | 6.79 | 8 | 3.48 | 14 |
| 15. improved | 2.71 | 17 | 5.33 | 9 | 3.92 | 11 | 4.15 | 11 | 7.42 | 7 | 3.76 | 13 |
| 25. great GP | 3.42 | 13 | 4.58 | 11 | 3.77 | 13 | 4.06 | 12 | 5.51 | 10 | 3.86 | 11 |
| 12. saying thanks | 2.59 | 19 | 3.64 | 16 | 3.52 | 15 | 4.03 | 13 | 5.35 | 11 | 2.76 | 17 |
| 22. disappointing | -4.97 | 8 | -7.57 | 3 | -1.15 | 31 | -2.14 | 26 | -5.23 | 12 | -2.82 | 16 |
| 41. the best ever | 2.62 | 18 | 3.46 | 19 | 2.86 | 18 | 3.29 | 16 | 4.14 | 18 | 2.52 | 19 |
| 43. recommend | 2.83 | 16 | 3.90 | 15 | 2.41 | 21 | 2.43 | 23 | 3.89 | 21 | 2.15 | 20 |
| 5. great vaccine help | 1.94 | 22 | 3.14 | 20 | 2.66 | 19 | 2.99 | 18 | 4.61 | 15 | 2.01 | 25 |
| 19. male relative went | 1.87 | 23 | 2.71 | 23 | 2.31 | 23 | 2.60 | 20 | 4.10 | 20 | 2.13 | 21 |
| 6. satisfied | 2.29 | 20 | 2.93 | 21 | 2.32 | 22 | 2.21 | 25 | 3.71 | 22 | 2.10 | 22 |
| 21. big changes | -1.39 | 28 | -2.63 | 25 | -1.66 | 24 | -2.45 | 21 | -4.15 | 17 | -2.62 | 18 |
| 28. effective help | 1.56 | 25 | 2.39 | 27 | 2.46 | 20 | 3.00 | 17 | 4.12 | 19 | 1.95 | 26 |
| 33. long wait | -1.67 | 24 | -3.52 | 18 | -1.18 | 28 | -1.74 | 30 | -2.57 | 28 | -0.92 | 34 |
| 32. long wait times | -1.37 | 29 | -1.24 | 36 | -2.90 | 17 | -1.83 | 28 | -1.84 | 35 | -2.08 | 23 |
| 46. poor experience | -0.66 | 36 | -0.33 | 46 | -3.46 | 16 | -2.78 | 19 | -2.45 | 32 | -2.03 | 24 |
| 44. long-term happy | 1.42 | 27 | 2.14 | 28 | 1.09 | 32 | 1.36 | 33 | 2.49 | 30 | 0.94 | 33 |
| 39. professional | 1.33 | 30 | 2.10 | 29 | 1.17 | 30 | 1.24 | 35 | 2.73 | 25 | 0.83 | 36 |
| 23. son treated | -0.66 | 37 | -0.65 | 40 | -1.49 | 25 | -2.44 | 22 | -1.69 | 36 | -1.37 | 27 |
| 38. friendly | 0.71 | 34 | 1.21 | 37 | 1.01 | 33 | 1.46 | 31 | 3.13 | 23 | 1.24 | 29 |
| 36. parking problem | 1.50 | 26 | 1.88 | 31 | 1.36 | 27 | 1.00 | 38 | 2.57 | 27 | 0.40 | 42 |
| 48. diabetes check | -0.05 | 44 | -0.46 | 42 | -1.17 | 29 | -2.05 | 27 | -1.85 | 34 | -2.89 | 15 |
| 24. [no meaning] | 1.17 | 31 | 2.06 | 30 | 1.00 | 34 | 1.30 | 34 | 2.52 | 29 | 0.90 | 35 |
| 42. impossible appointment | -1.99 | 21 | -3.93 | 14 | | | | | -2.67 | 26 | -0.44 | 40 |
| 14. booking roulette | -6.34 | 3 | -1.38 | 35 | -0.23 | 49 | | | -1.07 | 42 | -1.28 | 28 |
| 30. long-term condition | -0.35 | 41 | -2.55 | 26 | -0.56 | 42 | -1.82 | 29 | -3.10 | 24 | | |
| 29. no appointment | -0.54 | 39 | -2.63 | 24 | -0.60 | 40 | -0.45 | 44 | -1.40 | 37 | -1.09 | 30 |
| 16. give dignity | 0.65 | 38 | 1.07 | 38 | 0.45 | 43 | 0.92 | 39 | 1.93 | 33 | 0.96 | 31 |

| | | | | | | | | | | |
|---|---|---|---|---|---|---|---|---|---|---|
| 17. don't listen | -0.82 | 32 | -0.55 | 41 | -0.84 | 37 | -0.79 | 41 | -0.34 | 47 | -0.94 | 32 |
| 50. out of hours care | 0.18 | 43 | | | 1.37 | 26 | 1.38 | 32 | 1.09 | 40 | 0.11 | 47 |
| 35. hospital referral | | | | | -0.42 | 45 | -2.26 | 24 | -1.20 | 38 | -0.83 | 37 |
| 40. repeat prescription | 0.48 | 40 | 1.52 | 34 | 0.85 | 36 | 0.76 | 42 | 1.07 | 41 | 0.19 | 45 |
| 18. long-term experience | 0.80 | 33 | 1.67 | 33 | | | | | 1.11 | 39 | 0.32 | 44 |
| 1. distress phone booking | | | | | -0.88 | 35 | -0.10 | 46 | | | -0.71 | 38 |
| 11. hard appointments | -0.33 | 42 | -1.80 | 32 | 0.34 | 47 | 0.55 | 43 | | | | |
| 13. poor mental care | | | | | -0.81 | 38 | -1.01 | 37 | -0.21 | 49 | | |
| 8. bad opinions | | | | | | | 0.86 | 40 | 0.99 | 43 | 0.43 | 41 |
| 31. blood test | | | | | -0.02 | 50 | -1.23 | 36 | -0.13 | 50 | -0.61 | 39 |
| 57. [meaning not certain] | | | -0.03 | 47 | -0.80 | 39 | -0.35 | 45 | | | -0.34 | 43 |
| 7. can't choose doctor | 0.69 | 35 | 0.33 | 45 | | | | | 0.90 | 44 | 0.04 | 48 |
| 52. empathy | | | -1.02 | 39 | -0.44 | 44 | | | -0.82 | 45 | | |
| 10. decent practice | | | -0.34 | 44 | 0.58 | 41 | 0.04 | 47 | | | | |
| 45. time delay | | | -0.35 | 43 | 0.36 | 46 | | | -0.39 | 46 | -0.02 | 49 |
| 34. [meaning not certain] | -0.04 | 45 | | | | | | | 0.24 | 48 | -0.12 | 46 |
| 3. bad facilities | | | | | -0.24 | 48 | -0.04 | 48 | | | | |
| 51. pros and cons | | | | | | | | | | | | |
| 53. demand pressure | | | | | | | | | | | | |
| 56. lack common sense | | | | | | | | | | | | |

**Table A10: 30-topic LDA – Predictors for lasso models where star ratings are the dependent variable**

| | Model 1 | Model 2 | Model 3 | Model 4 | Model 5 | Model 6 |
|---|---|---|---|---|---|---|
| Topic 1 | 0 | 0.3911432 | 0 | 0 | 0 | 0 |
| Topic 2 | -9.076767 | -2.014217 | -3.706465 | -3.877787 | -2.099544 | -6.043043 |
| Topic 3 | 0 | 1.404052 | 0.0156617 | 2.726781 | 2.748896 | 1.79117 |
| Topic 4 | 2.83673 | 4.534612 | 6.825908 | 7.963718 | 9.33016 | 6.578943 |
| Topic 5 | 2.096279 | 3.995727 | 4.232543 | 4.966572 | 6.335657 | 3.39579 |
| Topic 6 | 4.977868 | 5.416236 | 5.007242 | 6.565092 | 7.936614 | 5.607379 |
| Topic 7 | 3.977873 | 5.138038 | 4.43614 | 4.492146 | 5.521623 | 3.75656 |
| Topic 8 | -6.124173 | -4.039576 | -0.313403 | -0.080392 | -3.233042 | -1.026911 |
| Topic 9 | -5.942907 | -6.737787 | -13.84749 | -11.93198 | -13.13358 | -7.90778 |
| Topic 10 | -1.587277 | -2.895708 | -3.796833 | -7.024889 | -6.846389 | -4.777935 |
| Topic 11 | 0 | -0.433927 | -0.019353 | -0.928606 | -0.193687 | -0.112656 |
| Topic 12 | 0 | 0.2901643 | 0.356878 | 0 | 0.0028051 | 0 |
| Topic 13 | 1.82196 | 2.999637 | 2.93703 | 4.309834 | 4.610383 | 3.075765 |
| Topic 14 | 5.038216 | 6.988456 | 6.771876 | 8.040449 | 9.754755 | 6.287765 |
| Topic 15 | -2.538325 | -4.255157 | 0 | 0 | -2.89584 | -1.381885 |
| Topic 16 | 0 | 0 | -1.02932 | 1.194027 | 0 | 0 |
| Topic 17 | -0.754918 | -0.156985 | -0.551399 | -0.399668 | -1.058872 | -1.220059 |
| Topic 18 | -0.461679 | -0.618314 | -3.425972 | -1.065881 | -1.685483 | -1.149781 |
| Topic 19 | 0 | 0 | 0 | 0 | 0 | 0 |
| Topic 20 | -3.777447 | -3.491196 | -2.65027 | -1.812122 | -3.549579 | -3.020006 |
| Topic 21 | 0 | 0 | 0 | 0 | 0 | 0 |
| Topic 22 | 1.096128 | 2.011053 | 2.235762 | 2.062248 | 2.983118 | 1.421168 |
| Topic 23 | 1.151248 | 1.80683 | 1.14825 | 1.309405 | 1.665083 | 1.08503 |
| Topic 24 | -0.398097 | -2.039163 | -0.121184 | -0.498358 | -2.434955 | -0.595727 |
| Topic 25 | -0.166396 | -1.757265 | -0.630698 | 0 | -0.744746 | -0.58713 |
| Topic 26 | 0.2331518 | 0.9428468 | 1.483587 | 1.823342 | 1.5774 | 1.160041 |
| Topic 27 | 2.421954 | 3.36193 | 3.70565 | 4.720741 | 5.4103 | 3.95075 |
| Topic 28 | -0.645797 | -7.313017 | -3.710486 | 0 | -9.231851 | 0 |
| Topic 29 | -6.997629 | -6.904233 | -12.51796 | -5.804027 | -5.712854 | -7.950538 |
| Topic 30 | -0.372418 | 0 | 0.0392828 | 0.5931976 | 0.4777672 | 0 |

# Random Forest model quality

Calculating the average of averages that we use in the paper: precision 0.39; recall 0.47; f1 0.36. Overall number of reviews is 146,388. At the disaggregate level, precision, recall and F1 scores for predicting the level of user satisfaction (number of review stars) is provided for each dimension of satisfaction below:

| phone access ease | | | |
|---|---|---|---|
| | precision | recall | f1score |
| 1 star | 0.622 | 0.420 | 0.546 |
| 2 star | 0.184 | 0.287 | 0.267 |
| 3 star | 0.214 | 0.284 | 0.297 |
| 4 star | 0.109 | 0.309 | 0.178 |
| 5 star | 0.888 | 0.626 | 0.629 |

| given dignity & respect | | | |
|---|---|---|---|
| | precision | recall | f1score |
| 1 star | 0.774 | 0.472 | 0.697 |
| 2 star | 0.037 | 0.260 | 0.070 |
| 3 star | 0.248 | 0.299 | 0.356 |
| 4 star | 0.084 | 0.309 | 0.149 |
| 5 star | 0.937 | 0.811 | 0.756 |

| likely to recommend | | | |
|---|---|---|---|
| | precision | recall | f1score |
| 1 star | 0.941 | 0.720 | 0.848 |
| 2 star | 0.006 | 0.803 | 0.012 |
| 3 star | 0.005 | 0.732 | 0.009 |
| 4 star | 0.009 | 0.712 | 0.017 |
| 5 star | 0.941 | 0.821 | 0.848 |

| appointment ease | | | |
|---|---|---|---|
| | precision | recall | f1score |
| 1 star | 0.925 | 0.570 | 0.679 |
| 2 star | 0.040 | 0.292 | 0.075 |
| 3 star | 0.029 | 0.332 | 0.054 |
| 4 star | 0.149 | 0.370 | 0.234 |
| 5 star | 0.838 | 0.545 | 0.654 |

| involved in care decisions | | | |
|---|---|---|---|
| | precision | recall | f1score |
| 1 star | 0.840 | 0.445 | 0.713 |
| 2 star | 0.013 | 0.331 | 0.025 |
| 3 star | 0.082 | 0.275 | 0.144 |
| 4 star | 0.059 | 0.286 | 0.107 |
| 5 star | 0.926 | 0.778 | 0.743 |

| up-to-date GP information | | | |
|---|---|---|---|
| | precision | recall | f1score |
| 1 star | 0.783 | 0.395 | 0.681 |
| 2 star | 0.012 | 0.370 | 0.023 |
| 3 star | 0.066 | 0.240 | 0.119 |
| 4 star | 0.107 | 0.279 | 0.182 |
| 5 star | 0.917 | 0.782 | 0.728 |

Random Forest model accuracy:

| | accuracy |
|---|---|
| phone access ease | 0.487 |
| appointment ease | 0.537 |
| given dignity & respect | 0.634 |
| involved in care decisions | 0.620 |
| likely to recommend | 0.772 |
| up-to-date GP information | 0.603 |

Confusion matrices (rows - star predictions, columns - star values). Matrix diagonal contains counts of correct predictions, precision and recall weights calculated on the column and row margins respectively:

| phone access ease | | | | | |
|---|---|---|---|---|---|
| | 1 | 2 | 3 | 4 | 5 |
| 1 | 14328 | 3363 | 3081 | 1050 | 1230 |
| 2 | 9160 | 3630 | 3682 | 1612 | 1694 |
| 3 | 6182 | 3035 | 4645 | 2439 | 5401 |
| 4 | 3275 | 1950 | 3463 | 3358 | 18768 |
| 5 | 1165 | 660 | 1507 | 2396 | 45314 |

| appointment ease | | | | | |
|---|---|---|---|---|---|
| | 1 | 2 | 3 | 4 | 5 |
| 1 | 39078 | 694 | 258 | 769 | 1441 |
| 2 | 15447 | 759 | 247 | 885 | 1553 |
| 3 | 7531 | 574 | 454 | 1855 | 5435 |
| 4 | 4098 | 390 | 248 | 4318 | 19874 |
| 5 | 2352 | 182 | 162 | 3845 | 33939 |

| given dignity & respect | | | | | |
|---|---|---|---|---|---|
| | 1 | 2 | 3 | 4 | 5 |
| 1 | 20755 | 555 | 2817 | 479 | 2192 |
| 2 | 9651 | 570 | 2686 | 607 | 1783 |
| 3 | 8136 | 618 | 4377 | 1207 | 3331 |
| 4 | 3520 | 312 | 3219 | 1386 | 8023 |
| 5 | 1935 | 140 | 1537 | 812 | 65740 |

| involved in care decisions | | | | | |
|---|---|---|---|---|---|
| | 1 | 2 | 3 | 4 | 5 |
| 1 | 24753 | 117 | 1049 | 669 | 2890 |
| 2 | 9821 | 166 | 705 | 499 | 1940 |
| 3 | 10312 | 103 | 1310 | 955 | 3390 |
| 4 | 7036 | 69 | 1089 | 1124 | 9876 |
| 5 | 3737 | 46 | 616 | 689 | 63427 |

| likely to recommend | | | | | |
|---|---|---|---|---|---|
| | 1 | 2 | 3 | 4 | 5 |
| 1 | 51384 | 11 | 4 | 5 | 3207 |
| 2 | 8455 | 57 | 1 | 2 | 1035 |
| 3 | 4919 | 0 | 30 | 6 | 1618 |
| 4 | 2824 | 2 | 3 | 89 | 7491 |
| 5 | 3801 | 1 | 3 | 23 | 61417 |

| up-to-date GP information | | | | | |
|---|---|---|---|---|---|
| | 1 | 2 | 3 | 4 | 5 |
| 1 | 19701 | 82 | 997 | 1657 | 2738 |
| 2 | 7924 | 137 | 724 | 1233 | 1727 |
| 3 | 9893 | 73 | 1067 | 1788 | 3362 |
| 4 | 8517 | 57 | 1126 | 2392 | 10270 |
| 5 | 3870 | 21 | 528 | 1496 | 65008 |

# Variety of LDA topic structures in reviews

GP reviews were grouped into 100 clusters to portray general patterns in how service users explain their experience. K-means clustering was used to generate the clusters because the method can efficiently process relatively high numbers of data points. Each node in Figure A3 represents a cluster or reviews. The biggest cluster has 3820 reviews and the smallest cluster has 230 reviews. 10 largest clusters in Figure A3 were labelled according to the most representative words in reviews included in respective clusters, and all clusters were coloured according to the dominant proportion of topics: red (negative), blue (positive) and green (neutral).

**Figure A3: Map of reviews clustered according to the proportions of LDA topics they contain**

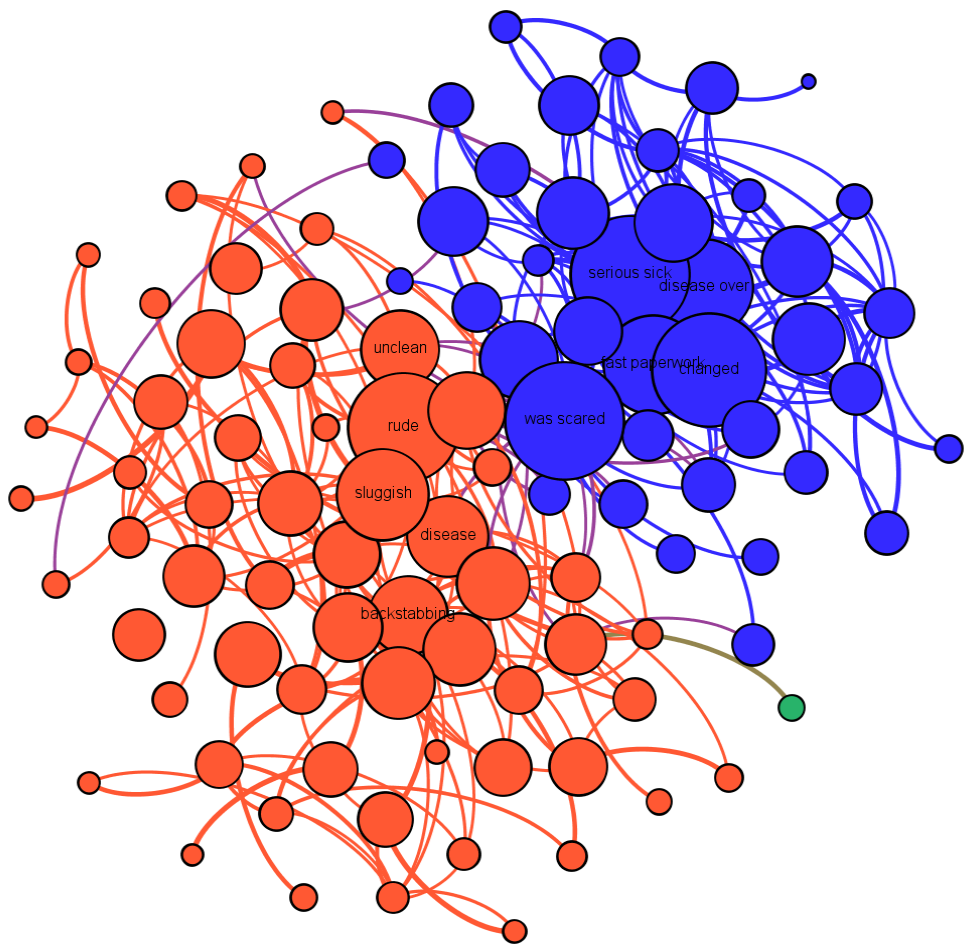

*Notes: (1) The graph, generated with Atlas method in Gephi v0.91, illustrates similarities of 146,388 reviews according to their topical structure. (2) Red review clusters indicate predominantly negative topics, blue ones indicate predominantly positive topics and the green cluster indicates predominantly*

*neutral topics. (3) Reviews are grouped into 100 clusters with k-means clustering method implemented in Python programming language. Positions of clusters were calculated based on pairwise Hellinger distance. The higher the Hellinger distance, the weaker the gravitational pull of two reviews. (3) 10 largest clusters of reviews were labelled according to the 10 most representative words in each cluster (measured with TF-IDF).*

K-means clustering begins with picking cluster centers (centroids) in the same space as data points. In this case, the algorithm was started with 100 cluster centre positions which were as far as possible from one another in a 57-dimensional space of reviews (the 57 dimensions were the topic proportions in individual reviews). Data points are matched to their nearest initial cluster centre (with Euclidean distance), and clusters' inertia is calculated. Inertia is a sum of squared distances of the data points to their respective cluster centres. After assignment of data points to their initial cluster centroids, new centroid locations are calculated by averaging values of all data points which make up the cluster. Then, having new centroid positions, all data points are again assigned to their nearest centroid and inertia is calculated, to be followed by another round of calculation of centroids. The process continues iteratively until reduction in the inertia score is not happening any more, for up to 100 iterations of the centroid calculations. K-means was used for cluster position optimisation with 10 different sets of initial starting cluster points, and the set of clusters with the lowest inertia was chosen for visualisation.

K-means clustering has two major weaknesses which are relevant to the task. First, it requires a manual choice of the number of clusters to generate. That's why 100 clusters were generated. Second, k-means works based on an assumption that data points which form clusters are normally distributed in all dimensions. For example, if data points occurred in ball-like clusters in a two-dimensional space, k-means would work well for clustering them. On the other hand, if data points formed spiral-shaped or anyway otherwise not ball-like patterns, k-means would likely misclassify some of the data points. Those 2 limitations were not deemed critical because the purpose of clustering was to carry out data reduction to visualize the most general patterns in topic composition. If there is a need to further improve the visualisation, it can be done with DBSCAN clustering. DBSCAN clustering does not require manual setting of the number of reviews to generate and can cope with irregular shapes of clusters to be generated. It is much more robust but with some additional computational cost.